\documentclass{ieeeaccess}

\usepackage{cite}
\usepackage{amsmath,amssymb,amsfonts}
\usepackage{algorithmic}
\usepackage{graphicx}
\usepackage{textcomp}
\usepackage{tabularx}
\usepackage{url}
\usepackage{hyperref}

\def\BibTeX{{\rm B\kern-.05em{\sc i\kern-.025em b}\kern-.08em
    T\kern-.1667em\lower.7ex\hbox{E}\kern-.125emX}}

\newcolumntype{L}{>{\raggedright\arraybackslash}X}
\newcolumntype{s}{>{\hsize=.5\hsize}X}
\newcolumntype{a}{>{\hsize=0.45\hsize}L}
\newcolumntype{d}{>{\hsize=0.55\hsize}X}
\begin{document}

\title{Explainable Predictive Maintenance: A Survey \\of Current Methods, Challenges and Opportunities}
\author{\uppercase{Logan Cummins}\authorrefmark{1}, 
\uppercase{Alex Sommers}\authorrefmark{1}, 
\uppercase{Somayeh Bakhtiari Ramezani}\authorrefmark{1}, 
\\
\uppercase{Sudip Mittal}\authorrefmark{1}, 
\uppercase{Joseph Jabour}\authorrefmark{2},
\uppercase{Maria Seale}\authorrefmark{2}
and 
\uppercase{Shahram Rahimi}\authorrefmark{1}
}

\address[1]{Department of Computer Science and Engineering, Mississippi State University, Mississippi State, MS 39762 USA }
\address[2]{U.S. Army Engineer Research and Development Center (ERDC), Vicksburg, MS, 39180, MS}

\tfootnote{This work by the Mississippi State University was financially supported by the U.S. Department of Defense (DoD) High Performance Computing Modernization Program, through the US Army Engineering Research and Development Center (ERDC) (\#W912HZ21C0014). The views and conclusions contained herein are those of the authors and should not be interpreted as necessarily representing the official policies or endorsements, either expressed or implied, of the U.S. Army ERDC or the U.S. DoD. Authors would also like to thank Mississippi State University's Predictive Analytics and Technology Integration (PATENT) Laboratory for its support.}

\markboth
{Cummins \headeretal: Explainable Predictive Maintenance: A Survey of Current Methods, Challenges and Opportunities}
{Cummins \headeretal: Explainable Predictive Maintenance: A Survey of Current Methods, Challenges and Opportunities}

\corresp{Corresponding author: Logan Cummins (e-mail: nlc123@cavs.msstate.edu).}

\begin{abstract}
Predictive maintenance is a well studied collection of techniques that aims to prolong the life of a mechanical system by using artificial intelligence and machine learning to predict the optimal time to perform maintenance. The methods allow maintainers of systems and hardware to reduce financial and time costs of upkeep. As these methods are adopted for more serious and potentially life-threatening applications, the human operators need trust the predictive system. This attracts the field of Explainable AI (XAI) to introduce explainability and interpretability into the predictive system. XAI brings methods to the field of predictive maintenance that can amplify trust in the users while maintaining well-performing systems. This survey on explainable predictive maintenance (XPM) discusses and presents the current methods of XAI as applied to predictive maintenance while following the \emph{Preferred Reporting Items for Systematic Reviews and Meta-Analyses} (PRISMA) 2020 guidelines. We categorize the different XPM methods into groups that follow the XAI literature. Additionally, we include current challenges and a discussion on future research directions in XPM.
\end{abstract}

\begin{keywords}
eXplainable Artificial Intelligence, XAI, Predictive Maintenance, Industry 4.0, Industry 5.0, Interpretable Machine Learning, PRISMA
\end{keywords}

\titlepgskip=-21pt

\maketitle
\section{Introduction} \label{sec:introduction}
The history of technological advancements within the past couple of hundred years is well documented. These centuries and decades can be categorized into what is described as revolutions, i.e. Industrial Revolutions \cite{ramezani2023scalability}. The most recent of these is agreed to be known as the fourth industrial revolution or Industry 4.0 \cite{ramezani2023scalability, Leng2022-ab, Longo2020-zl, Nahavandi2019-ht}. 

Industry 4.0 is categorized by bridging the gap between machinery through hardware and software connectivity \cite{Lavopa2021WhatPlatform}. This revolution is characterized by the inclusion of human-machine interfaces, AI, and internet of things technologies \cite{Lavopa2021WhatPlatform}. Through these technologies, we can become more automated and efficient with new challenges that come with big data and cyber-physical systems. One of the problems created from this revolution has centered around the optimization of mechanical systems. 

One method of optimizing mechanical systems is to minimize the downtime the system may suffer from due to break-downs and repairs. To tackle this level of optimization, researchers of Industry 4.0 have developed the field of predictive maintenance (PdM). PdM encompasses many different problems in the field of maintenance, but an overarching representation of PdM involves monitoring the system as it is in the present and alerting for any potential problems such as a specific anomaly or time until failure \cite{ramezani2023scalability, cummins2021deep}. While this problem that exists in the cyber-physical realm has been well studied from the perspective of deep learning models, statistical models, and more, the people that get impacted by these systems have had considerably less attention. This change of focus leads us into the fifth industrial revolution or Industry 5.0. 

While the mechanical systems were the focus of the fourth industrial revolution, human-centered challenges have become the focus of the fifth revolution. As described by Leng et al. \cite{Leng2022-ab}, humans must be important in the processes related to these important decision-making systems. Nahavandi et al. \cite{Nahavandi2019-ht} illustrates Industry 5.0 in the realm of a factory line. The human performs a task that is assisted by an artificial intelligent agent that can increase the productivity of the human. 

As these systems are moving the focus away from mechanics and towards humans, a different area must be brought to the forefront. The way to address human-centered processes can be derived from the fields of eXplainable AI (XAI) and Interpretable Machine Learning (iML). XAI and iML are extensively researched from multiple fields on a wide array of problems including the various problems in PdM. Our article's main contribution involves using the \emph{Preferred Reporting Items for Systematic Reviews and Meta-Analyses} statement to organize the XAI and iML works applied to PdM. We also describe and categorize the different methods, note challenges found in PdM and provide key aspects to keep the field of Explainable Predictive Maintenance (XPM) moving forward. 

The article is organized in the following manner. In Section \ref{sec:background}, important information surrounding explainability, Interpretable Machine Learning, and predictive maintenance are described. Section \ref{sec:search} describes the literature search performed including identification, screening, and inclusion. In Sections \ref{sec:results},\ref{sec:xai_pdm} and \ref{sec:iml_pdm}, the results of the literature review are categorized and discussed in detail. Section \ref{sec:challenges} discusses challenges in the field that remain to be addressed, and Section \ref{sec:conclusion} provides our closing remarks.

\section{Background} \label{sec:background}
To accommodate readers of varying backgrounds, we briefly explain a couple of key topics needed for understanding the importance of this research, namely Explainable Artificial Intelligence (XAI), Interpretable Machine Learning (iML), and Predictive Maintenance (PdM). We will also discuss the distinction between XAI and iML to inform the readers of the perspective with which we evaluated the literature.
\subsection{Explainability and Interpretability in Artificial Intelligence}
\label{sec:explainability}
The fine distinction between \emph{explainability} and \emph{interpretability} in the context of AI and ML has raised considerable debate \cite{speith2022review}. While several researchers argue that the terms are synonymous, viewing them as interchangeable to simplify discussions \cite{zhou2021evaluating, nauta2022anecdotal, rong2022towards, Sharma2022-tr}, others assert that they capture distinct concepts \cite{marcinkevivcs2023interpretable, clinciu2019survey, kim2023help, Neupane2022-ly, Nor2021-lh, Ali2023-av, rojat2021explainable, sokol2021explainability}. Interestingly, a third perspective points out that one term is a subset of the other, adding another layer to the discourse \cite{schoonderwoerd2021human, lopes2022xai, Vollert2021-tw}.

To ensure clarity and coherence in this article, we consider that \emph{explainability} and \emph{interpretability} are related yet distinct. While there exists a certain degree of overlap, they emphasize different facets of machine learning. 
\subsubsection{Explainable Artificial Intelligence}
\begin{figure}[ht]
    \centering
    \includegraphics[width=0.5\textwidth]{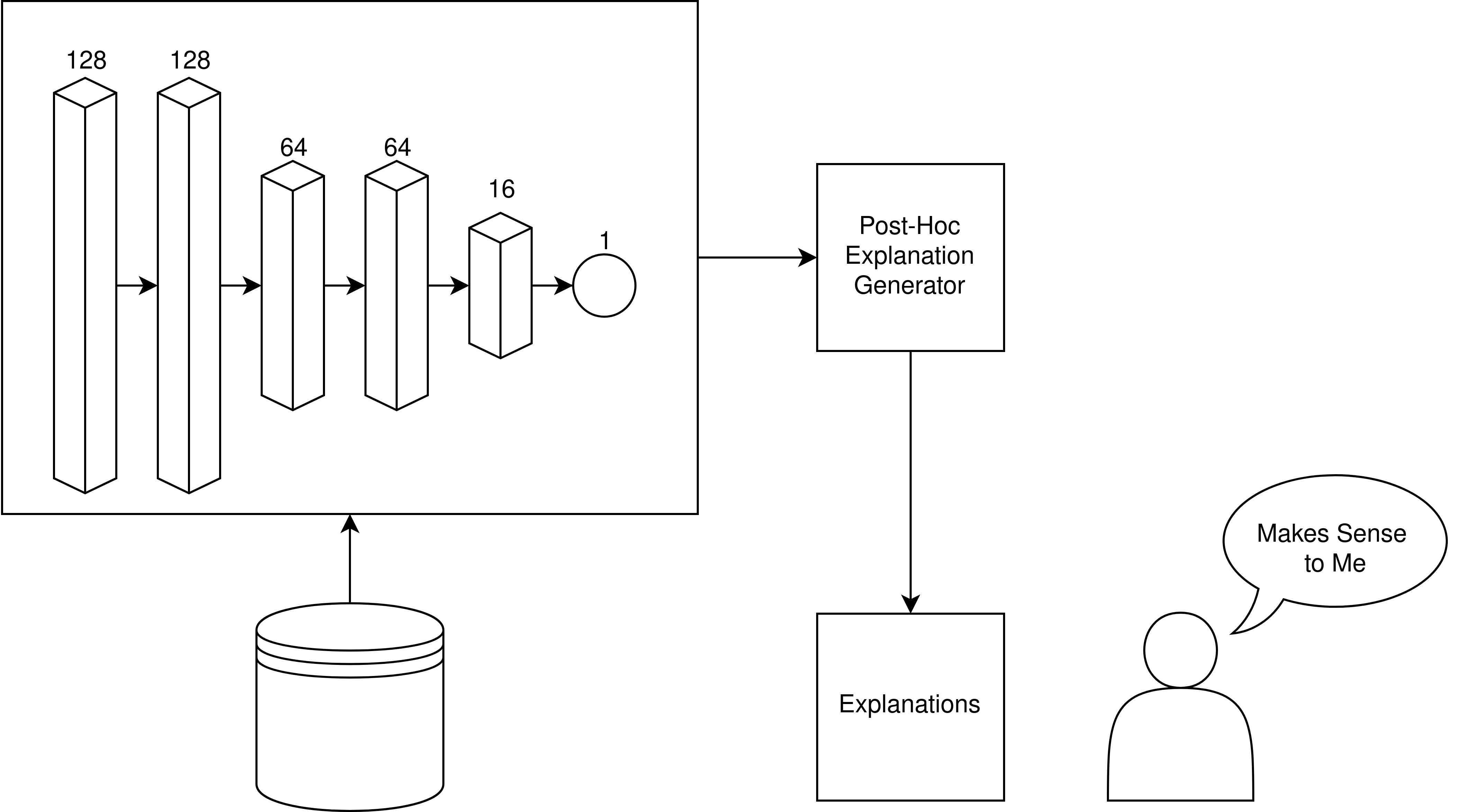}
    \caption{Visualization of XAI Design Cycle}
    \label{fig:xai}
\end{figure}

The rapidly growing field of eXplainable Artificial Intelligence (XAI) aims to demystify AI systems by clarifying their reasoning mechanisms and subsequent outputs \cite{speith2022review}. XAI methodologies can typically be classified based on features such as the scope of explanation—whether global or local—and the techniques employed for generating explanations, like feature perturbation. A unifying theme across these methods is the endeavor to interpret the workings of an already-trained model. As Sokol et al. succinctly put it, \emph{explainability is for the model's output} \cite{sokol2021explainability}. From a more analytical standpoint, XAI predominantly encompasses post-hoc strategies to shed light on otherwise opaque, black-box models \cite{Nor2021-lh}. This paradigm is illustrated in Figure \ref{fig:xai}, where a model's explanations are constructed to enhance user comprehension. 

\paragraph{Model-Agnostic and Model-Specific.} Explainable methods can be categorized based on their suitability for addressing various types of black-box models. Methods that are applicable to models regardless of their architecture are called \emph{model-agnostic.} Common methods that fall into this category are Shapley Additive Explanations (SHAP) \cite{Lundberg2017-eu} and Local Interpretable Model-agnostic Explanations (LIME) \cite{Ribeiro2016-zc}. These methods and additional model-agnostic methods are described in Section \ref{model-agnostic}. The opposite of these methods are known as \emph{model-specific.} Model-specific methods such as Class Activation Mapping (CAM) \cite{Zhou2016-gv} for Convolutional Neural Networks (CNNs) are designed to take advantage of the architecture already to provide explainability. These methods and others are described in Section \ref{model_specific}.

\paragraph{Local Explanations and Global Explanations.} Another way of classifying explainable methods is by the scope of the explaination. These scopes are commonly described as either \emph{local} or \emph{global.} Local explanations aim at explaining the model's behavior for a single data point. Global explanations provide reasoning that represents the model's behavior for any data point. 

\paragraph{XAI Example}
To give a concrete example of XAI, a researcher may want to use a Long Short-Term Memory neural network  for time-series analysis due to its temporal modeling capabilities \cite{cummins2021deep, ramezani2023scalability}. Common deep learning models like this one are not commonly interpretable, so to make it explainable, the researcher might consider using a simpler model, i.e., linear regression, decision tree, etc., to serve as a surrogate for post-hoc explanations. These explanations would then be presented to the user/developer/stakeholder to better explain the behavior of an inherent black-box architecture.

\subsubsection{Interpretable Machine Learning}
\begin{figure}[ht]
    \centering
    \includegraphics[width=0.5\textwidth]{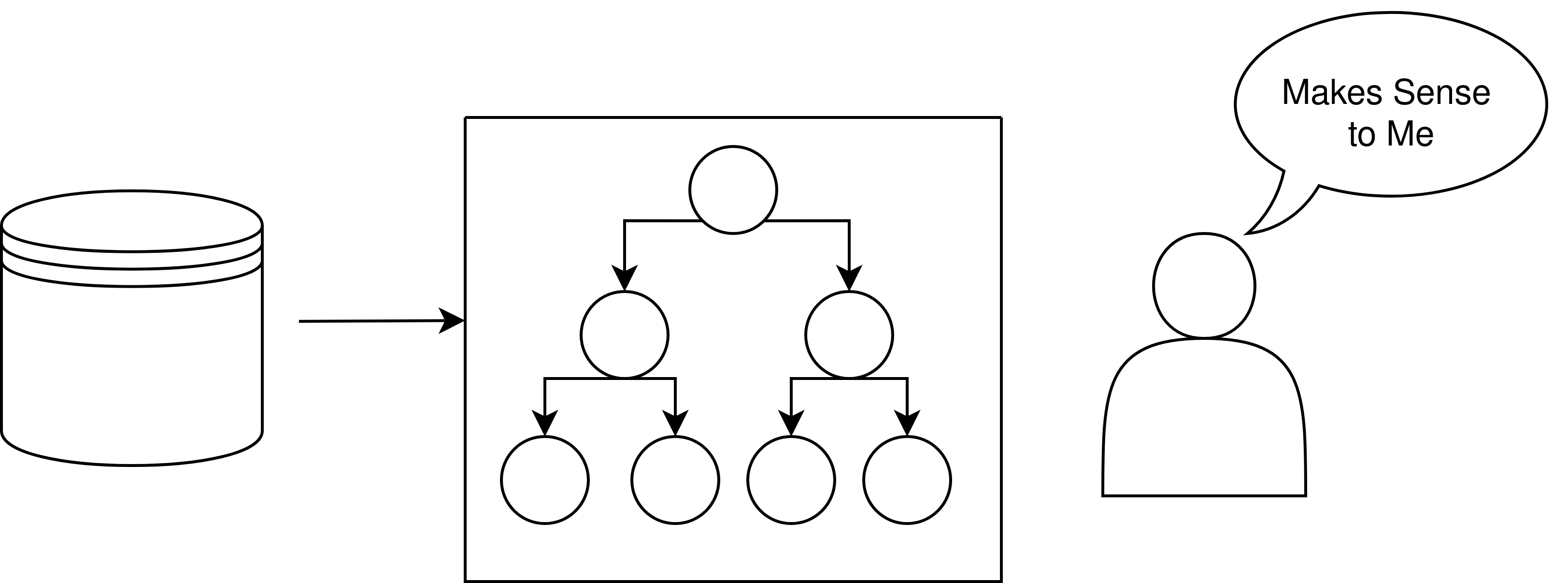}
    \caption{Visualization of interpretable ML Design Cycle}
    \label{fig:iml}
\end{figure}

Interpretable Machine Learning (iML) describes ML models that are referred to as \emph{white-} or \emph{gray-boxes} \cite{marcinkevivcs2023interpretable}, and their interpretability is enforced by architectural or functional constraints. Between the two, architectural constraints make models simple enough to understand, while physical constraints attempt to cast the model's computations in terms of real-world features. While XAI focuses on the model's output, iML focuses on the model itself \cite{sokol2021explainability}. This has also been stated as \emph{intrinsic interpretability} as to separate it from \emph{post-hoc explainability} methods \cite{Vollert2021-tw, molnar2020interpretable}. As follows, this article will equate iML with models that are intrinsically interpretable through methods of structural constraints, physical bindings, etc. This can be seen in Figure \ref{fig:iml}, where there is no need for translating the model through an explainable method. 

For a concrete example, a researcher may have a problem that could benefit from a simple logistic regression classifier. With such a simple architecture, the network itself would be interpretable as it would be clear what inputs affect what outputs. One could also extrapolate the overarching equation if the network is simple enough. This illustrates inherent interpretability.
\subsection{Predictive Maintenance}
\label{sec:pm}
Predictive maintenance (PdM) is a subcategory of prognostics and health management (PHM) that has seen widespread attention in recent years \cite{ramezani2021similarity, ramezani2023scalability, Wen2022-re, Vollert2021-tw}. PdM utilizes AI and previous failure information from mechanical systems to predict a fault or downtime in the future \cite{cummins2021deep, ramezani2021hmm, ramezani2023scalability}. PdM is implemented with a variety of tools, including anomaly detection, fault diagnosis and prognosis \cite{Wen2022-re, Vollert2021-tw}.

Anomaly detection and fault diagnosis have a very distinct difference. Whereas anomaly detection aims at determining whether a fault occured or not, fault diagnosis aims to identify the cause of a fault \cite{tsui2015prognostics, Wen2022-re}. This means that anomaly detection can be thought of as a binary classification problem, and fault diagnosis can be thought of as an extension of anomaly detection to a multi-classification problem. Finally, prognosis deals with predicting the remaining useful life (RUL) or time until failure \cite{Wen2022-re, cummins2021deep, ramezani2023scalability}. This puts prognosis in the domain of regression problems. Now that these terms are defined and categorized into their different problems, we can discuss the PRISMA compliant systematic search that we performed.

\section{Systematic Search}
\label{sec:search}
\begin{figure}[!tbp]
  \centering
  \includegraphics[width=0.5\textwidth]{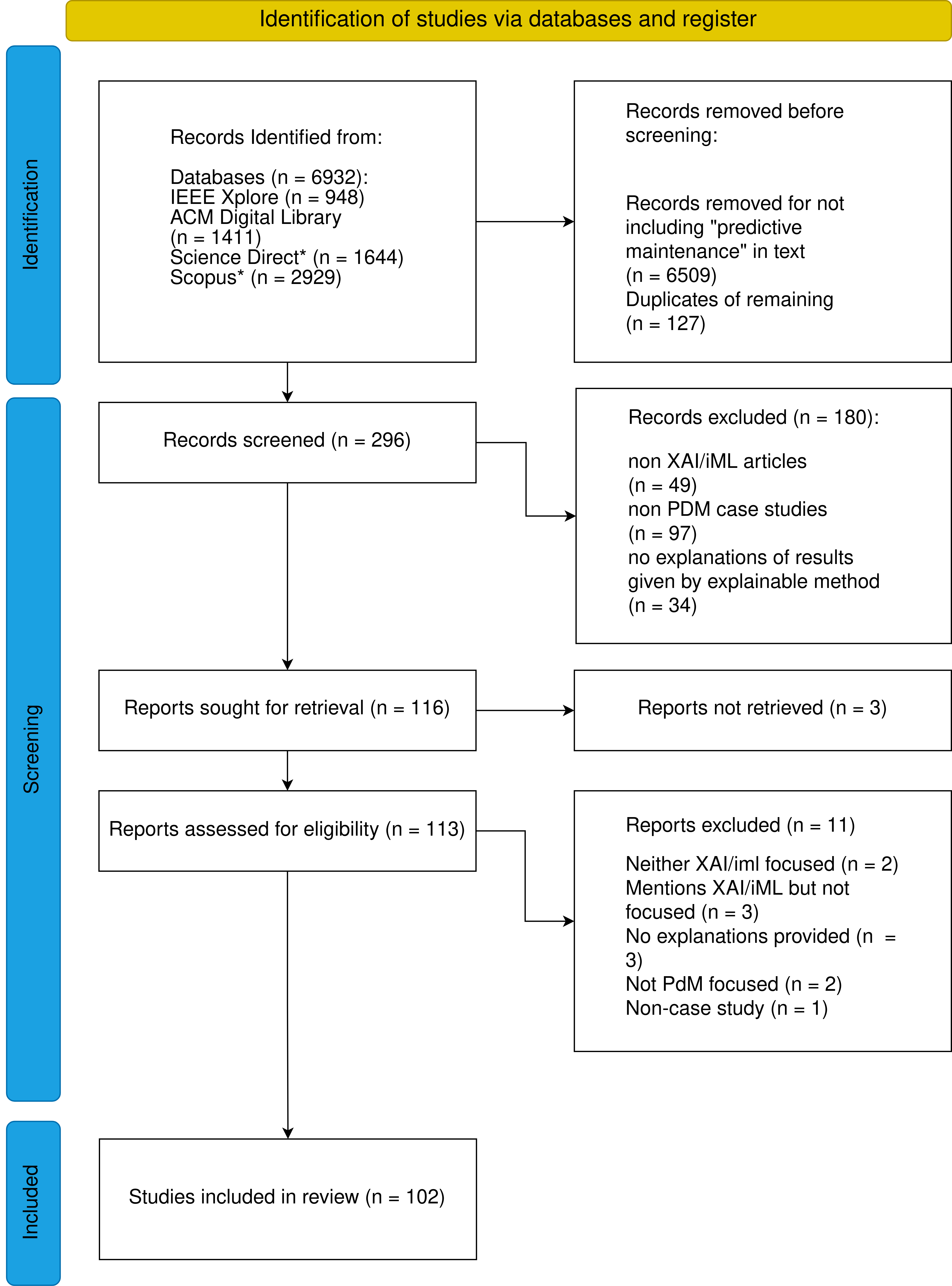}
    \caption{PRISMA Search}
    \label{fig:prisma}
\end{figure}

We utilized the \emph{Preferred Reporting Items for Systematic Reviews and Meta-Analyses} (PRISMA) 2020 statement \cite{page2021prisma, haddaway2022prisma2020} to layout a systemized methodology of performing a literature review. The full process can be seen in Fig. \ref{fig:prisma}.

\subsection{Identification}
In identifying the potential databases, we focused on popular computer science publishers as well as general scientific publishers. We utilized the following databases for literature searches: IEEE Xplore, ACM Digital Library, ScienceDirect and Scopus, all of which were accessed on June 21, 2023. To capture as much as we could, we searched titles, keywords, and abstracts with two ideas in mind: \emph{XAI and iML} and \emph{PdM}. 

In the former case we used \emph{explainable OR interpretable OR xai} to capture the first grouping of papers. This should gather papers with common phrases like \emph{explainable artificial intelligence, explainable machine learning, interpretable ML, XAI, etc.} To capture the PdM aspect, we provided more explicit words so as to represent the research area better. We used \emph{prognos* OR diagnos* OR RUL OR remaining useful life OR predictive maintenance or detection.} This would capture ideas such as \emph{prognosis, prognostics, diagnosis, diagnostics, detection, etc.} 

In research, words like \emph{prognosis} and \emph{diagnosis} appear in medically related articles. This makes sense as many can attest that they would go to their physician for a diagnosis. To minimize the inclusion of medical literature, ScienceDirect and Scopus were set to look at Engineering and Computer Science related articles only. Even with this selection, the initial pool of research was 6932 articles. 

This narrowing down of papers was not as effective as we initially expected as only the titles, keywords, and abstracts were checked. Prior to removing duplicates, we also removed articles that did not mention \emph{predictive maintenance} inside of the article. After removing those papers and duplicates, the initial screening started with 296 articles. 
\subsection{Exclusion Criteria and Screening}
Our initial screenings involved skimming through the abstracts, main objectives, conclusions, and images of the articles. These initial screenings utilized the following exclusion criteria:
\begin{enumerate}
    \item Neither XAI nor iML are a main focus of the article.
    \item Articles are not PdM case studies.
    \item No explanation or interpretation is provided.
\end{enumerate}
The need for the first tow criteria is easily apparent. Many articles would mention one of the search terms from XAI/iML, but they would not fall into this category of work (n = 49). This would mainly emerge as using the words \emph{explainable} or \emph{interpretable} in a sentence of the abstract. Similarly, to the PdM case studies, many articles mention \emph{diagnosis}, and such, in a sentence without it being the focus of the article (n = 97). However, the third criterion needs a more in-depth explanation.

When stating that an architecture is interpretable or explainable, a certain expectation is implanted in the reader's mind. This applies to any concept whether it be computer science related or not. One of the expectations that we agreed upon was providing proof of interpretability or explainability. This would necessitate the explanation from the explainable method or the inherent interpretation of the interpretable model. With this expectation in mind, a few articles (n = 34) were removed before in-depth screening due to a mention of an explanatory method without any output of the said method. This finalized a screening population of 116 articles which were sought for retrieval. Three were not retrieved by our resources. Upon further examination, those three articles seem to lead to dead URLs.

For final assessment of eligibility, all of the resources were read. Many of the articles that were excluded were not available outside of a small preview. Of the remaining 113 articles, 11 were excluded for the following reasons:
\begin{itemize}
    \item Three mention XAI/iML in the abstract but do not utilize any methods that we could find.
    \item Two were neither XAI nor iML. These mention search terms in the abstracts, but do not build on them.
    \item Three offer no interpretations of their interpretive method.
    \item Two mention PdM in the abstract but do not focus on PdM in an experiment.
    \item One was not a case study.
\end{itemize}
\subsection{Inclusion}
After careful review of the articles, we finalized a population of 102 articles. Our findings and these articles are now discussed in Section \ref{sec:results}.

\section{Search Results}
\label{sec:results}
\begin{figure}[!tbp]
  \centering
  \includegraphics[width=0.5\textwidth]{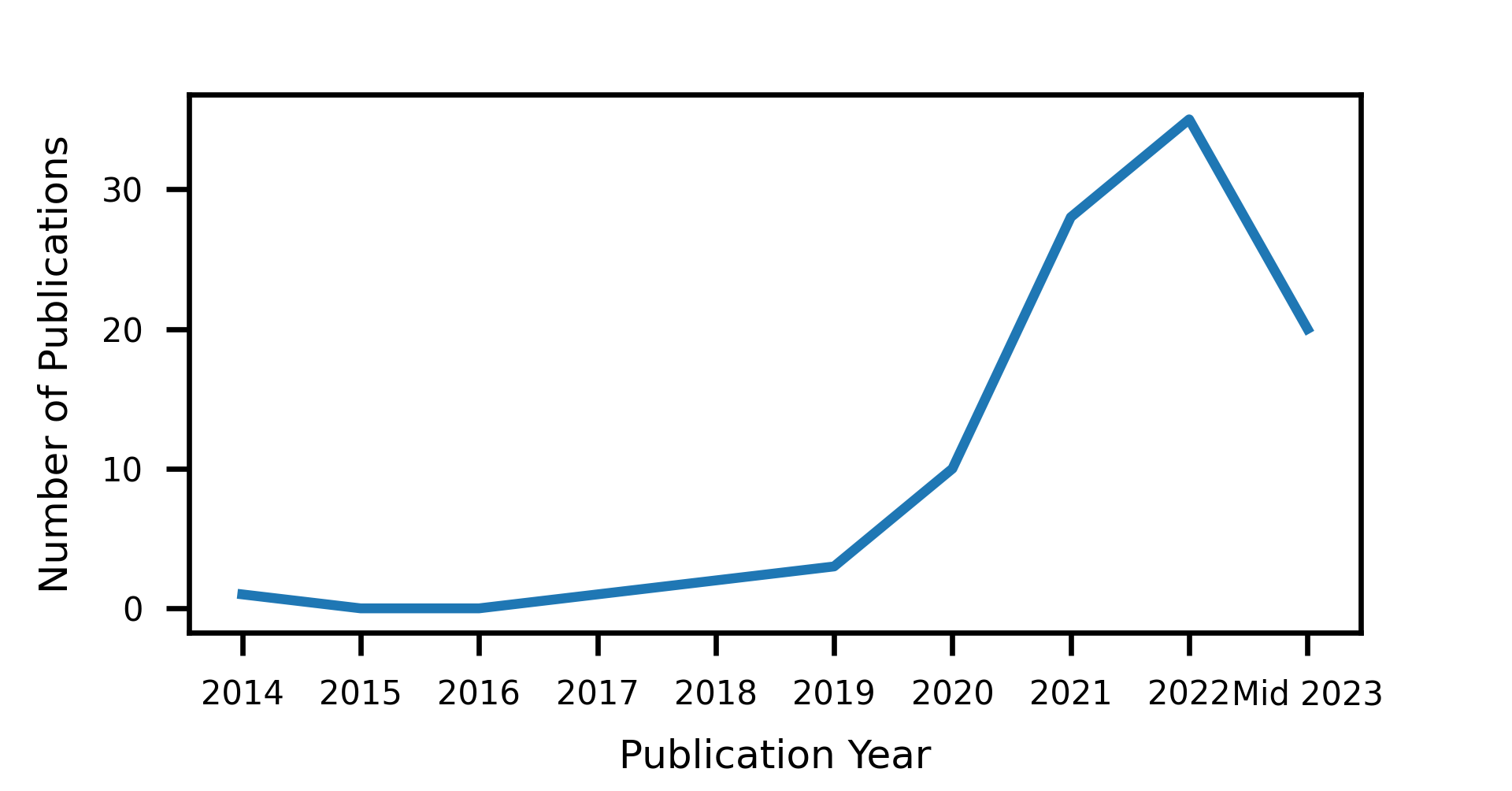}
    \caption{Articles published per year in our inclusion results}
    \label{fig:temporal}
\end{figure}

\begin{figure}[!tbp]
  \centering
  \includegraphics[width=0.5\textwidth]{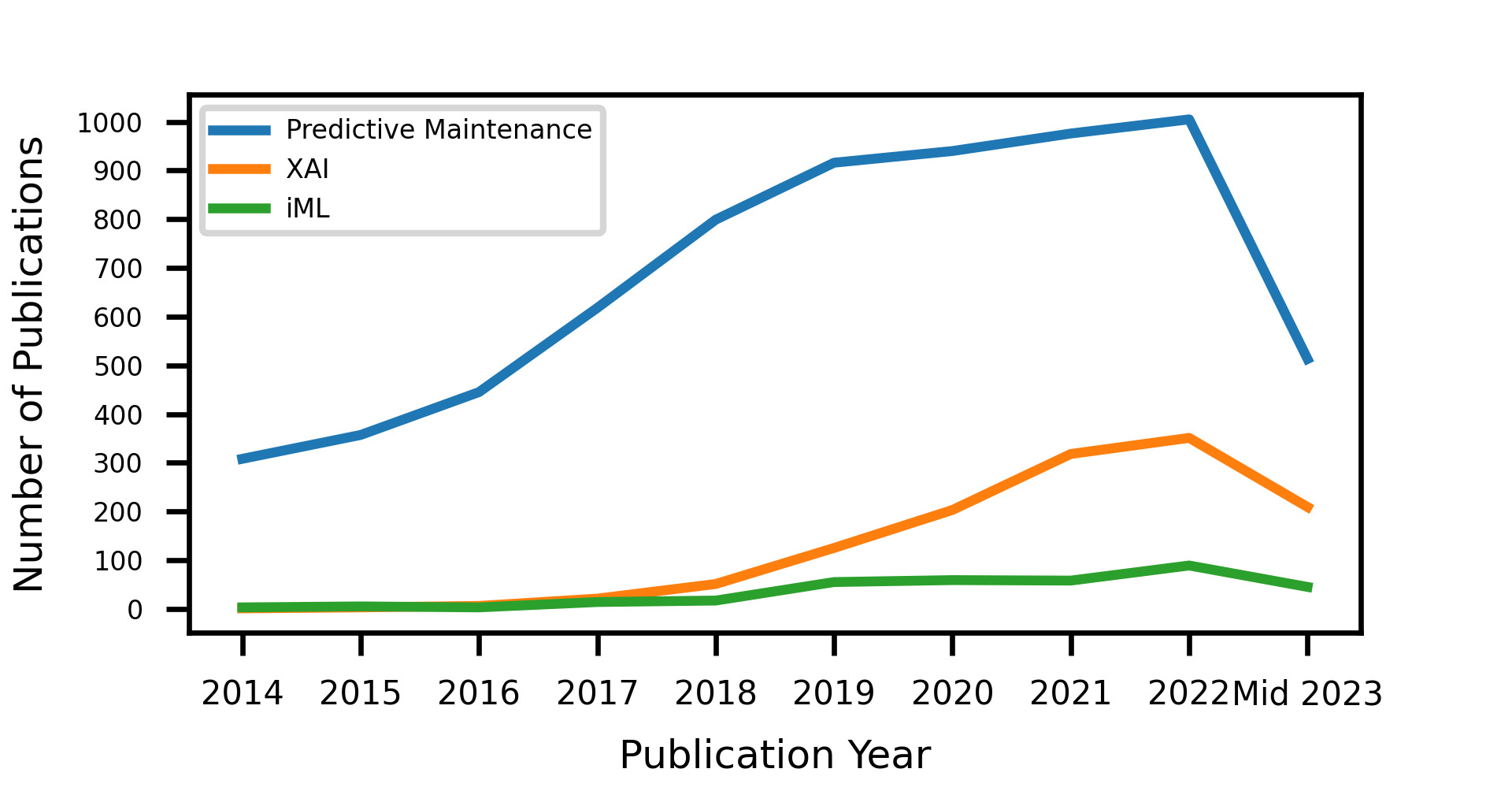}
    \caption{Google Search Trend for PdM, XAI, and iML from our article years}
    \label{fig:google}
\end{figure}

To paint an overarching picture of our results, Fig. \ref{fig:temporal} shows a break-down of our inclusion population grouped by year. This shows a clear increasing trajectory in publications that can be explained by a few potential factors. Firstly, the popularity of predictive maintenance continues to increase, as shown in \cite{ramezani2023scalability} and in Fig. \ref{fig:google}, as we move to a big-data centric world in industry. This provides more opportunities to implement these very large and very complex neural architectures for making important decisions. The importance of these decisions leads to a second reason for increasing importance, trust. 

Many articles discuss the importance of increasing the \emph{trust} of the users in the model while decreasing the \emph{bias} in black-box models  \cite{Ding2021-ez, manco2017fault, Protopapadakis2022-ye, khan2022explainable}. Rojat et al. define trust as achieved once a model can effectively explain its decisions to a person \cite{rojat2021explainable}. This would necessitate some sort of explainable or inherently interpretable architecture that could give the users insight. Furthermore, Vollert et al. \cite{Vollert2021-tw} even state that trust is a \emph{prerequisite} for a successful data-driven application. 

\begin{figure}[!tbp]
  \centering
  \includegraphics[width=0.5\textwidth, height=0.2\textheight]{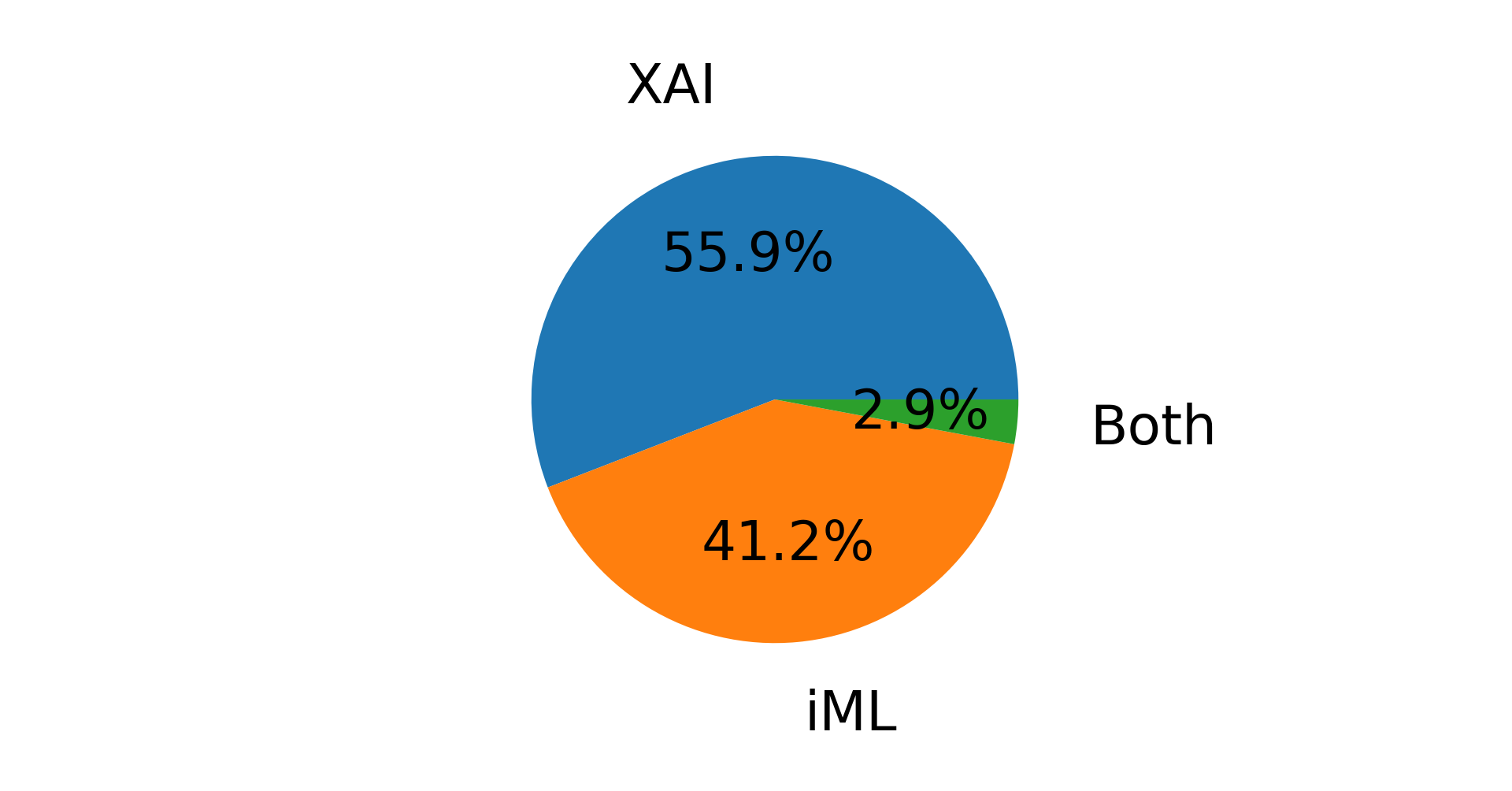}
    \caption{Distribution of XAI and iML in the search results}
    \label{fig:piechart}
\end{figure}

Looking at Fig. \ref{fig:piechart}, our findings reflect the idea that XAI is slightly more popular than iML in PdM. One potential reason could be the desire to make use of the benefits from complex models. Many of the articles utilize architectures such as Deep Convolutional Neural Networks \cite{Solis-Martin2023-po} or Long Short-term Memory Neural Networks \cite{Ferraro2023-jv} due to their high performance in the application. With the inherent black-box nature of these models, these researchers need post-hoc explainable methods. This desire for XAI over iML seems to be affecting specific PdM tasks more than others. 

\begin{figure}[!tbp]
  \centering
  \includegraphics[width=0.45\textwidth, height=0.2\textheight]{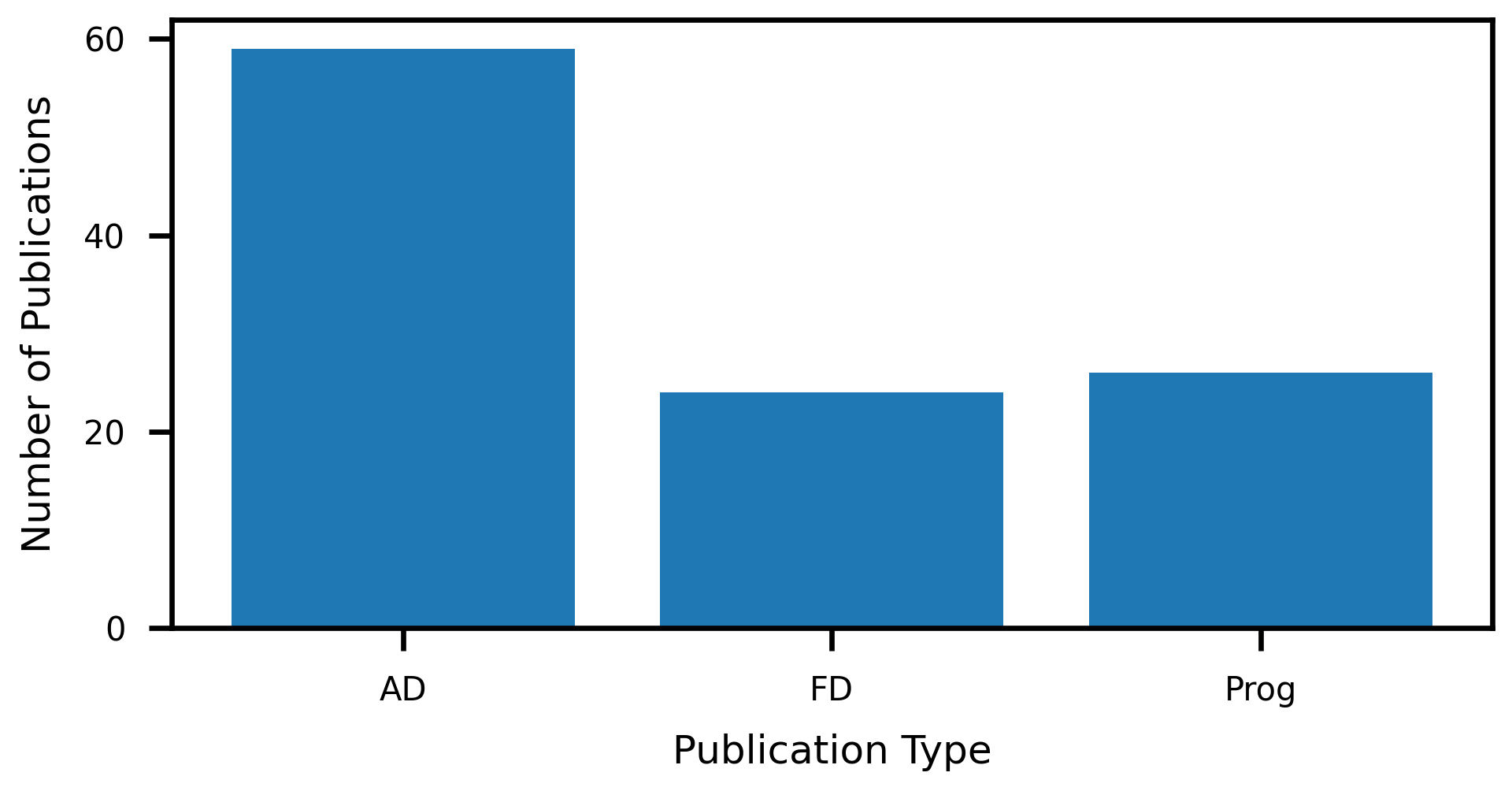}
    \caption{Papers per Anomaly Detection (AD), Fault Diagnosis (FD), and Prognosis (Prog)}
    \label{fig:paper_breakup}
\end{figure}

\begin{figure}[!tbp]
  \centering
  \includegraphics[width=0.45\textwidth, height=0.2\textheight]{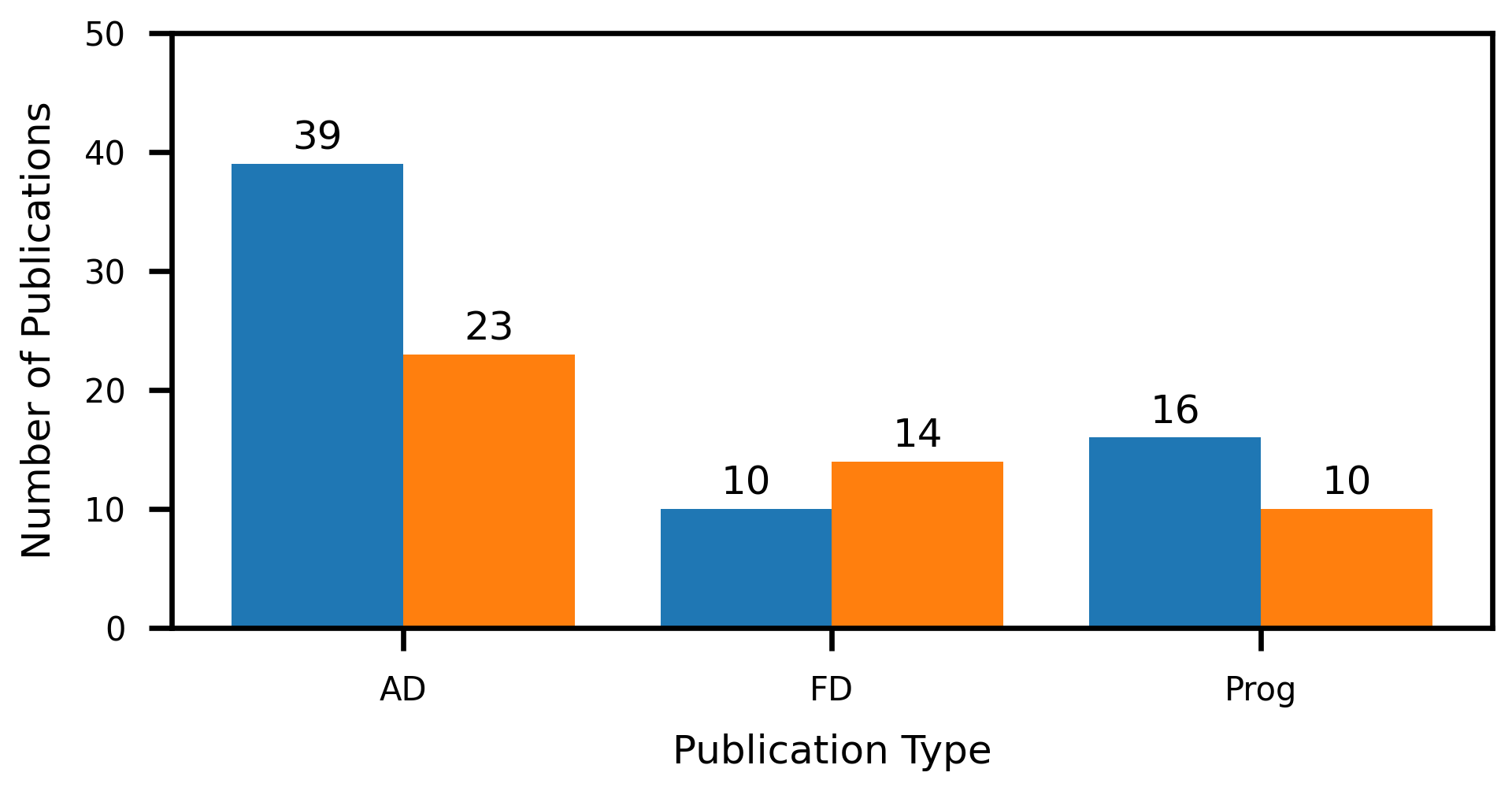}
    \caption{Split between XAI and iML per category of predictive maintenance}
    \label{fig:xai-iml-split}
\end{figure}

The articles are categorized according to PdM task in Fig. \ref{fig:paper_breakup}, and those are further distinguished into XAI and iML within tasks in Fig. \ref{fig:xai-iml-split}. Our article population reflects \emph{anomaly detection} as the main task that utilizes XAI and iML. Fault diagnosis and prognosis are virtually the same in number of articles published within this population; however, Fig. \ref{fig:xai-iml-split} shows that the interest in XAI and iML are reversed in these groups. Succinctly, prognosis focuses on XAI, while diagnosis focuses on iML. 
We now describe the many methods that were applied to the varying datasets seen in Table \ref{tab:datasets}. These methods are split between section \ref{sec:xai_pdm} for XAI methods and section \ref{sec:iml_pdm} for iML methods. Additionally, specific articles of interest can be found in Table \ref{tab:articles}. 

\begin{table}[ht]
\centering
\caption{Datasets from the Literature Search}
\begin{tabular}{| l | l |}
\hline
{\textbf{Datasets}}&{\textbf{Articles}}\\
\hline
Bearings and PRONOSTIA \cite{nectoux2012pronostia, QIU20061066}& \cite{yao2023integrated, brito2022explainable, yang2022noise, mey2022explainable, li2022waveletkernelnet, pu2023restricted, xin2022fault, BRITO2023120860, ben2022end, sanakkayala2022explainable, serradilla2020interpreting, kothamasu2007adaptive, lughofer2022transfer}\\
& \cite{Solis-Martin2023-po, Alfeo2022-or, wang2020online, Ding2021-ez} \\
Vehicles or vehicle subsystem & \cite{wang2021ifp, panda2023ml, xia2022fault, fan2022incorporating, li2023wavcapsnet, jang2019anomaly, ming2019interpretable, Oh2020explainable, hafeez2022dtcencoder, ribeiro2022online, Li2021-ri, voronov2021forest, Han2022-yk}\\
CMAPSS \cite{saxena2008turbofan} & \cite{brunello2023monitors, wu2018k, khan2022explainable, jakubowski2021anomaly, brunello2020learning, costa2022variational, sayed2022explainable, Jakubowski2022-hh, Kononov2023-ys, Solis-Martin2023-po, Protopapadakis2022-ye, Jing2022-yq, Waghen2022-na}\\
 & \cite{Abbas2022-mm}\\
General Machine Faults and Failures \cite{brito2002using} & \cite{brito2022explainable, glock2021explaining, BRITO2023120860, matzka2020explainable, torcianti2021explainable, Remil2021-kr, Ghasemkhani2023-rd}\\
Trains & \cite{liu2023causal, trilla2023unsupervised, ERRANDONEA2023103781, steenwinckel2021flags, li2014improving, Allah_Bukhsh2019-bk, manco2017fault}\\
Gearboxes \cite{cao2018preprocessing} & \cite{brito2022explainable, li2022waveletkernelnet, BRITO2023120860, Hajgato2022-ia}\\
Artificial Dataset & \cite{jakubowski2021explainable, mylonas2022local, mey2022explainable}\\
Hot or Cold Rolling Steel & \cite{jakubowski2021explainable, jakubowski2022roll, jakubowski2021anomaly}\\
Mechanical Pump & \cite{xu2022physics, salido2004comparison, langone2020interpretable}\\
Aircraft & \cite{janakiraman2018explaining, kothamasu2007adaptive}\\
Amusement Park Rides & \cite{berno2021machine, anello2022anomaly}\\
Particle Accelerators & \cite{marcato2021machine, felsberger2020explainable}\\
Chemical plant & \cite{bellini2021deep, choi2022explainable} \\
Maritime & \cite{kim2021explainable, michalowska2021anomaly, Bakdi2022-gw}\\
Semi-conductors \cite{misc_secom_179} & \cite{gashi2023impact, cao2020using}\\
Air Conditioners & \cite{wang2021ifp}\\
Hard Drives \cite{Klein_2021} & \cite{brunello2023monitors, Ferraro2023-jv, Amram2021-gt}\\
Tennessee Eastman Process \cite{katser2021unsupervised} & \cite{brunello2023monitors}\\
Compacting Machines & \cite{anello2022anomaly}\\
UCI Machine Learning Repository \cite{Dua:2019} & \cite{scott2023computational} \\
Wind Turbines \cite{hansen_vasiljevic_sørensen_2021} & \cite{roelofs2021autoencoder, Beretta2021-gk, beretta2021improved}\\
Transducers & \cite{upasane2021big}\\
Lithium-ion Batteries \cite{attia2020closed} & \cite{csalodi2021mixture, Solis-Martin2023-po, Wang2023-px}\\
Heaters & \cite{verkuil2022automated}\\
Computer Numerical Control data & \cite{lorenti2022cuad}\\
Textiles & \cite{ugli2022uzadl}\\
Plastic Extruders & \cite{lourencco2022using}\\
Press Machine & \cite{serradilla2021adaptable}\\
Coal Machinery & \cite{hermansa2021sensor}\\
Refrigerators & \cite{xu2021deep}\\
Gas Compressors & \cite{steurtewagen2021adding}\\
Hydraulic Systems & \cite{keleko2023health}\\
Iron Making Furnaces & \cite{chen2020temporal}\\
Cutting Tools & \cite{Schmetz2021-pf} \\
Power Lines \cite{vsb-power-line-fault-detection} & \cite{Simmons2021-hj} \\
Communication Equipment & \cite{Zhang2021-ee}\\
Water Pump & \cite{Upasane2023-al}\\
Oil Drilling Equipment & \cite{Xia2023-bg}\\
Solenoid operated valves & \cite{Tod2023-cf} \\
Coal Conveyors & \cite{Mahmoodian2022-zk}\\
Temperature Monitoring Devices & \cite{konovalenko2022generating}\\
Distillation Unit & \cite{Dhaou2021-qx}\\
Water Pipes \cite{data.wa.gov.au} & \cite{Castle2020-pc}\\
\hline
\end{tabular}
\label{tab:datasets}
\end{table}

\begin{table}[ht]
\centering
\caption{Explainable Methods from the Literature}
\begin{tabularx}{0.5\textwidth}{| a | d |}
\hline
{\textbf{Method}}&{\textbf{Articles}}\\
\hline
Shapley Values (\ref{shap})& \cite{bellini2021deep,choi2022explainable, hermansa2021sensor, jakubowski2021anomaly, jakubowski2021explainable, kim2021explainable, lourencco2022using} \cite{khan2022explainable, sayed2022explainable, gashi2023impact, serradilla2021adaptable, steurtewagen2021adding, brito2022explainable, Jakubowski2022-hh, Kononov2023-ys,Solis-Martin2023-po, Ferraro2023-jv, Li2021-ri} \\
LIME (\ref{lime})& \cite{Ferraro2023-jv, khan2022explainable, Li2021-ri,Jakubowski2022-hh,Solis-Martin2023-po,serradilla2020interpreting,Alfeo2022-or,jang2019anomaly,Protopapadakis2022-ye, sanakkayala2022explainable,torcianti2021explainable,mey2022explainable}\\
Feature Importance (\ref{feature_importance}) & \cite{Allah_Bukhsh2019-bk,Bakdi2022-gw,manco2017fault,Schmetz2021-pf,voronov2021forest,Simmons2021-hj,Alfeo2022-or, marcato2021machine,Remil2021-kr, Ghasemkhani2023-rd}\\
LRP (\ref{LRP}) & \cite{felsberger2020explainable,Han2022-yk,mey2022explainable,Wang2023-px,Solis-Martin2023-po}\\
Rule-based (\ref{rule_based_explainer}) & \cite{wu2018k,brunello2023monitors,ribeiro2022online,brunello2020learning}\\
CAM and GradCAM (\ref{cam}) & \cite{Solis-Martin2023-po,Oh2020explainable,BRITO2023120860,mey2022explainable}\\
Surrogate (\ref{surrogate}) & \cite{Zhang2021-ee,jakubowski2021explainable,ERRANDONEA2023103781,glock2021explaining}\\
Visualization (\ref{visualization}) & \cite{ugli2022uzadl, michalowska2021anomaly, xin2022fault, costa2022variational}\\
DIFFI (\ref{diffi}) & \cite{berno2021machine,lorenti2022cuad,brito2022explainable}\\
Integrated Gradients (\ref{integrated_gradient}) & \cite{Hajgato2022-ia, serradilla2021adaptable}\\
Causal Inference (\ref{causal}) & \cite{trilla2023unsupervised}\\
ACME (\ref{acme}) & \cite{anello2022anomaly}\\
Statistics (\ref{statistics}) & \cite{fan2022incorporating}\\
SmoothGrad (\ref{smootgrad})& \cite{serradilla2021adaptable}\\
Counterfactuals (\ref{counterfactual}) & \cite{jakubowski2022roll}\\
LionForests (\ref{lionforest}) & \cite{mylonas2022local}\\
ELI5 (\ref{eli5}) & \cite{serradilla2020interpreting}\\
Saliency Maps (\ref{salience}) & \cite{Solis-Martin2023-po}\\
ARCANA (\ref{arcana}) & \cite{roelofs2021autoencoder}\\
\hline
\end{tabularx}
\label{tab:explainable}
\end{table}
\section{Explainable AI in Predictive Maintenance}\label{sec:xai_pdm}
XAI in predictive maintenance captures a wide range of methods that can be categorized in several ways. To not repeat information, the methods are broken up into three subsections: model-agnostic, model-specific, and combination.  

\subsection{Model-Agnostic}\label{model-agnostic}
This section describes the explainable methods in our population, seen in Table \ref{tab:explainable}, that could be applied to any architecture. These methods are colloquially known as \emph{model-agnostic} explainable methods \cite{Belle2021-nd}. These methods found in this section can be applied to any architecture and consist of SHAP in Section \ref{shap}, LIME in Section \ref{lime} and additional related methods.

\subsubsection{Shapley Additive Explanations (SHAP).}\label{shap} SHAP values were introduced by Lundberge et al. as a unified measure of feature importance \cite{Lundberg2017-eu}. SHAP is based on three properties that are shared with classical Shapley value estimation: local accuracy, missingness, and consistency. Local accuracy refers to the ability of the simplified input to \emph{at least} match the output of the input from the data. Missingness refers to the features that are missing from the simplified input. Succinctly, this states that if a feature is not useful to the explanation, then it is not useful to the model. Finally, consistency brings the idea that the importance of a feature should stay the same or increase regardless of the other features.

By far, SHAP is the most used method seen in our sample. Moreover, SHAP is one of the few methods that has been applied to the problems of anomaly detection\cite{bellini2021deep, choi2022explainable, hermansa2021sensor, kim2021explainable, jakubowski2021anomaly}, fault diagnosis \cite{lourencco2022using, steurtewagen2021adding}, and prognosis \cite{sayed2022explainable, Kononov2023-ys, gashi2023impact}. This is likely due to its wide versatility as a model-agnostic method that can provide global explanations. 
 
Steurtewagen et al. \cite{steurtewagen2021adding} created a framework for fault diagnosis that consists of three parts: data collection, prognosis, and diagnosis. Importantly, in the data collection phase, they received the reports that were associated with the faults. The prognosis section used an XGBoost algorithm to detect a fault occurring. The diagnosis utilized SHAP to determine the features that are important to the output of XGBoost. These features are validated using the reports that accompany the fault. 

Choi et al. \cite{choi2022explainable} proposed a method for explainable unsupervised anomaly detection to predict system shutdowns for chemical processes. Their method consisted of what they call a period-independent framework and a period-integrated framework. The period-independent framework searched for the best anomaly detection model and applied the explainable method. In the period-integrated framework, they applied real-time information to the model chosen from the previous framework. They found that the isolation forest provided the best results in the period-independent framework based on the number of unplanned shutdowns detected, and they utilized show SHAP as an effective way of measuring root cause analysis.

Gashi et al. \cite{gashi2023impact} conducted predictive maintenance on a multi-component system. Their objective was to model interdependencies and assess the significance of the interdependencies. Prior to training their Random Forest model, they used visual exploration to study interdependencies. They used two methods to justify the use of interdependencies: statistics and XAI. They used chi-squared testing to show that the performance of a model with interdependencies is better (p < 0.001). When applying SHAP to the random forest, they showed that the interdependency variables were usually among the top explainer features. This adds validity to SHAP as an explainable method in terms of the accuracy of its explanations.

\subsubsection{Local Interpretable Model-agnostic Explanations (LIME).}\label{lime} LIME was introduced by Ribeiro et al. as a way of explaining any model using a local representation around the prediction \cite{Ribeiro2016-zc}. This is done by sampling around the given input data and training a linear model with the sampled data. In doing this, they can generate an explanation that is faithful to that prediction while using only information gained from the original model. 

Protopapadakis et al. \cite{Protopapadakis2022-ye} computed the RUL as applied to the CMAPSS turbofan dataset. They initially attempted to perform RUL prediction with two models, a random forest and a deep neural network. They found the random forest to perform poorly, which would lead to poor explanations. Their deep neural network achieved high performance, so they applied LIME. They compared two LIME explanations, one for early life and one for late life with a specific fault. They found that LIME was able to label the important features for failures that reflected the physical faults. Additionally, they showed that LIME would have a more difficult time labeling the important features when it was applied to segments with no faults as anything could occur in the future. 

Allah Bukhsh et al. \cite{Allah_Bukhsh2019-bk} discussed multiple tree-based classifiers for predicting the need for maintenance events, i.e., anomaly detection, for train switches. From their pool of tree-based classifiers, including decision tree, random forest, and gradient boosted tree, they identified gradient boosted tree as the most accurate amongst the models when predicting if a problem would occur. In a separate test, they had the same models predict specific types of anomalies. In this experiment, random forest outperformed the rest. For interpretability, they implemented LIME to learn from the outputs of the random forest. The researchers intend that the output from LIME will help establish trust in the model for domain experts and decision makers

\subsubsection{Feature Importance.}\label{feature_importance} Feature importance refers to the idea that some of the input features have more influence on the output than others. For example, when determining if an image is a dog, the background that has no pixels of the dog would potentially be less important than the pixels with the dog. Feature importance is typically assessed using techniques like SHAP and LIME, but various approaches exist in the literature.

Many researchers have applied different methods of feature importance calculations. Bakdi et al. \cite{Bakdi2022-gw} tackled predictive maintenance for ship propulsion systems. They combined balanced random forest models and multi-instance learning to achieve a high true positive rate which was then explained via Gini feature importance. Schmetz et al. \cite{Schmetz2021-pf} also applied Gini feature importance to verify a Tree Interpreter \cite{Saabas_2014} for their random forest classifier. 

Other researchers have ranked their features in different ways. Manco et al. \cite{manco2017fault} performed fault prediction to train systems where they ranked time steps by how anomalous they were within a time window. This ranking was performed by mixture modeling of the prior probability of the trend with the probability of the trend being normal behavior. Marcato et al. \cite{marcato2021machine} applied anomaly detection to particle accelerators where permutation-based feature importance to guide further model development. 

Finally, Voronov et al. \cite{voronov2021forest} and Ghasemkhani et al. \cite{Ghasemkhani2023-rd} each proposed different methods of calculating feature importance that tackle different problems. Voronov et al. proposed a forest-based variable selector called Variable Depth Distribution (VDD) that addressed the issue of variable interdependencies through clustering of features. The important features appeared in multiple clusters. Ghasemkhani et al. developed Balanced K-Star to deal with the imbalance problem commonly found in predictive maintenance. To add explainability, they applied chi-square to determine the important features in the machine failure.

\subsubsection{Layer-wise Relevance Propagation (LRP).}\label{LRP} LRP was introduced by Bach et al. \cite{Bach2015-yw} as an explainable method that assumes that a classifier can be decomposed into several layers of computation. LRP works with the concept of a relevance score that measures how important a feature is to an output. LRP works by extrapolating the relevance to the input layer by moving backwards through the architecture starting at the output layer. The importance of an input feature can then be measured as a summation of features it impacts through the architecture.

LRP falls into the category of model-agnostic which can be seen in the use-cases in the literature. Felsberger et al. \cite{felsberger2020explainable} applied LRP to multiple architectures including kNN, random forest, and CNN-based models. Through LRP, they found that the CNN architectures were learning important features which led to higher performance. Han et al. \cite{Han2022-yk} performed fault diagnosis for motors using the notable model LeNet \cite{lecun1998gradient}. Through the use of LRP, they were able to bring explainability to a notable architecture.

Wang et al. \cite{Wang2023-px} proposed a method of using explainability as a method of driving the training process. They utilized LRP to calculate feature importance for the training data. The importance calculations were embedded  for optimizing the model's performance. They introduced this explainability-driven approach to the problem of aging batteries, and showed its superb accuracy when compared to a data-driven approach.

\subsubsection{Rule-based Explainers.}\label{rule_based_explainer} Rule-based explainers use a combination of the black-box model and the training data to create a series of IF-THEN rules. These rules are generally created using combinatorial logic (ANDs, ORs, and NOTs) to combine the features in the IF portion of the rules. The THEN portion of the rules are populated by the result from the model, usually a class or a predicted value. The rules are then presented as explanations or may be used as a replacement for the black-box model itself. 

Even in rule-based explainers, there are numerous methods that have been used. Wu et al. \cite{wu2018k} proposed the K-PdM (KPI-oriented PDM) framework, a cluster-based HMM based on key performance indicators (KPIs). A KPI is a vector of one feature of fine-grained deterioration, and a combination of KPIs reflect the health of a machine. The health was modeled as an HMM for each KPI. These HMMs were converted into a rule-based reasoning system for explainability.

Brunello et al. \cite{brunello2020learning, brunello2023monitors} showed twice that temporal logic can be used in anomaly detection. Firstly, they showed that linear temporal logic could be added to an online system for monitoring failures \cite{brunello2020learning}. They again showed that temporal logic could be used in a different approach to the same problem. Brunello et al. \cite{brunello2023monitors} created syntax trees that utilized bounded signal temporal logic statement. The trees were altered using an evolutionary approach to predict failure in Blackblaze Hard Drive\cite{Klein_2021}, Tennessee Eastman Process\cite{katser2021unsupervised}, and CMAPSS\cite{saxena2008turbofan} datasets, commononly used datasets for PdM of hard drives, electrical processes and turbofans. This method led to great performance with rule-based explanations. 

Ribeiro et al. \cite{ribeiro2022online} applied XAI to the online learning process using a Long Short-term Memory AutoEncoder (LSTM-AE) for modeling public transport faults. Simultaneously, the authors' system learned regression rules that explained the outputs of the model. While their system was learning to map the anomalies, the output of their model was fed into Adaptive Model Rules (AMRules), a stream rule learning algorithm. They applied their method to four public transport datasets, and they output their global and local rule-based explanations given used in their system.

\subsubsection{Surrogate Models.}\label{surrogate} Surrogate models are simpler models that are used to represent more complex models. These surrogate models generally take the form of simple decision trees and linear/logistic regression models. The simplistic nature of these models makes them interpretable; however, their use has their interpretability as an explainable method for a black-box model. 

When utilizing a surrogate model as an explainable method, the surrogate model must be inherently interpretable as a way of allowing an explanation to be gathered from the main model. Glock et al. \cite{glock2021explaining} utilized two ARIMA models to explain a random forest model. One ARIMA model learned the same data as the random forest, and the second ARIMA model learned the residual errors from the random forest. While the random forest is not explainable, the two ARIMA models could show what the random forest could and could not learn.

Zhang et al. \cite{Zhang2021-ee} proposed an alarm-fault association rule extraction based on feature importance and decision trees. Their process started with a weighted-random forest. Feature selection was performed to gather the important features in the abnormal state. These features were used to create a series of C4.5 decision trees that model different features. Once their random forest was trained and predicted a fault, the decision tree with the highest accuracy could be used to extrapolate an explanation of the fault.

Errandonea et al. \cite{ERRANDONEA2023103781} tested XAI on edge computing with all possible models in H2O.ai's AutoML to perform their fault diagnosis. After determining the optimal architectures, they trained a decision tree surrogate model to add explainability to their autoML process. By optimizing hardware and accuracy, they showed that explainable predictive maintenance could theoretically occur on edge computing devices. 

\begin{figure*}[!tbp]
  \centering
  \includegraphics[width=\textwidth]{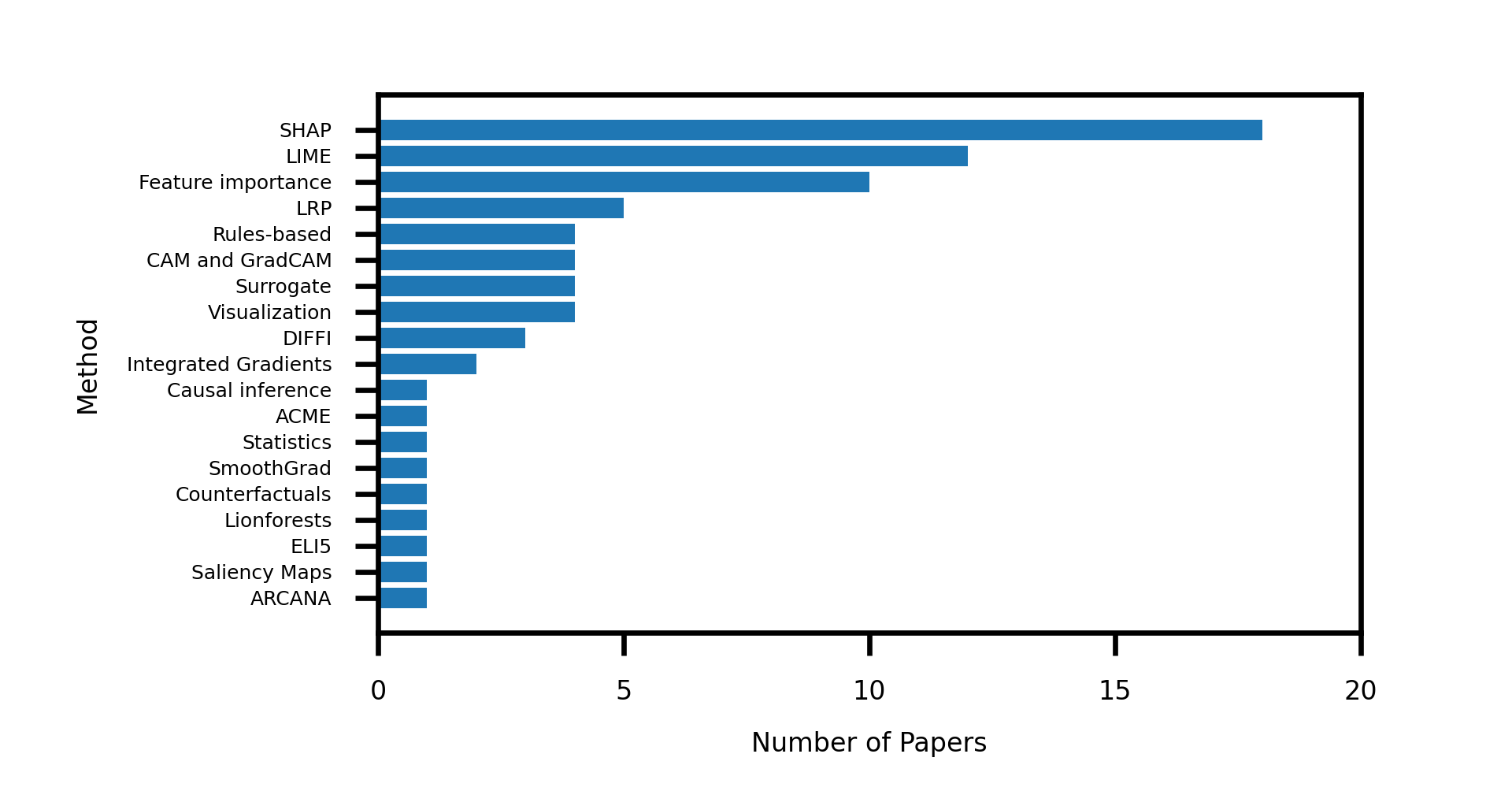}
    \caption{Use of XAI Methods}
    \label{fig:xaiMethods}
\end{figure*}

\subsubsection{Integrated Gradients.}\label{integrated_gradient} Integrated gradients was introduced by Sundararajan et al. \cite{Sundararajan2017-bv} to attribute the prediction of a deep architecture to its input features. They introduce two axioms, sensitivity and implementation invariance, to build their explainable method. Sensitivity is achieved if \emph{for every input and baseline that differ in one feature but have different predictions then the differing feature should be given a non-zero attribution.} Implementation invariance means \emph{attributions are always identical for two functionally equivalent networks.} With these axioms in mind, the integrated gradients are calculated via small summations through the layers' gradients.   

Hajgato et al. \cite{Hajgato2022-ia} introduced the PredMaX framework for predictive maintenance which identified sensitive machine parts and clustered time periods. It works in two steps: a deep convolutional autoencoder was applied to the data, and clustering was performed on the latent space in the autoencoder. From the clusters, they showed which channels contribute to the transition from normal to abnormal. Additionally, the integrated gradients technique was used to extract the relevant sensor channels for a malfunctioning machine part.

\subsubsection{Causal Inference.}\label{causal} Causality goes beyond the notion of statistics dependencies as it shows a true relationship between two or more variables \cite{Janzing2013-ml}. Causality can be measured in \emph{causal strength} which measures the change in distribution of n-1 variables when one variable has been changed \cite{Janzing2013-ml}. Causality is not an easy quality to analyze as it can only be truly discovered by repeated observations of a phenomenon occurring given an event; however, causal inference has been a method of XAI that some researchers have utilized. 

Trilla et al. \cite{trilla2023unsupervised} designed an anomaly detection framework based around a denoising variational autoencoder (VAE) and an MLP. They extracted intra-subsystem and inter-subsystem patterns by making the time series data into voxels. The VAE generalized the embeddings. Finally the MLP  was used to create a smooth diagnosis probabilistic function. They applied their method on a locomotion dataset and utilized causal inference via the Peter-Clark algorithm to answer the question "Did the VAE learn cause-effect relationships?" They found that the VAE could at best be described as modeling a correlation relationship, but this limitation was mainly attributed to limited data availability. 

\subsubsection{Visualization.}\label{visualization} Visualization techniques do not take any one specific form. Generally, these visualizations take the form of visualizing weights; however they may also take the form of visualizing specific examples. Whatever the case, these methods benefit the users by providing an image enlightens the user to the inner workings of the architecture. 

Visualizations can be utilized in many ways for explainability. Michalowska et al. \cite{michalowska2021anomaly} use visualizations to compare healthy and anomalous data. Costa et al.\cite{costa2022variational} utilized visualizations coupled with a recurrent variational encoder. They show that the latent space created by the encoder can add explainability. When input data with similar RULs pass through the encoder, they show the latent spaces are similar for those with similar RULs. 

Xin et al.\cite{xin2022fault} aimed to address bearing fault diagnosis via a novel model named logarithmic-short-time Fourier transform modified self-calibrated residual network (log-STFT-MSCResNet). The STFT extracts time-frequency features from raw signals to retain physical meaning of fault signatures which are visualized for explainability. The MSCResNet is used to enlarge the receptive field without introducing more parameters. With the combination of the two, they aim to have high accuracy even under unknown working conditions. They compared their model to popular models such as LSTM and ResNet18. log-STFT-MSCResNet performed among the best even under unknown working conditions, had a small number of features and had a shorter training time than the others.

\subsubsection{Accelerated Model-agnostic Explanations (ACME).}\label{acme} ACME was introduced by Dandolo et al. \cite{Dandolo2023-gy} as a method of quickly generating local and global feature importance measures based on perturbations of the data. For global explanations, they take a vector that holds the mean of each feature through the entire dataset; this is known as the baseline vector. Then a variable-quantile matrix is created that holds the different quantiles of the features. This matrix is used to gather predictions that would represent each quantile. The global feature importance is finally calculated for each feature by computing the standardized effect over each quantile. To get a local explanation, the baseline vector is replaced with the specific data point that is meant for explaining.

Anello et al. \cite{anello2022anomaly} applied ACME to the problem of anomaly detection to compare it to SHAP. They utilized isolation forest to detect anomalies as it is commonly used for detecting outliers or anomalies. An anomaly score was used as a label for the time series to represent the problem as a regression task which allows ACME to be applied. After applying SHAP and ACME to a roller coaster dataset and a compacting machine dataset, they found a drastic speed up by using ACME with all of the data while SHAP would be slower even with access to 30\% of the data.

\subsubsection{Statistics.}\label{statistics} As a method of explaination applied to the problem of predictive maintenance, statistical tests can be used to compare the distribution of the features between different classes. 

Fan et al. \cite{fan2022incorporating} developed ML methods that take advantage of physics knowledge for added interpretability. Their case study was fault detection of leak-related faults in vehicle air systems. They applied three physics equations to their data that would model the air leakage. Moreover, they used that data in the training data of their kNN and MLP models. Results showed that the physics-assisted models to outperform the non-assisted models.

\subsubsection{Smooth Gradients (SmoothGrad).}\label{smootgrad} SmoothGrad was developed by Smilkov et al. \cite{Smilkov2017-ap} to produce a gradient-based sensitivity map. The intuition behind SmoothGrad involves differentiating the predicting model with respect to the input. This derivative creates a sensitivity map that represents how much difference a change in each pixel of the input would make to the classification \cite{Smilkov2017-ap}. Moreover, this sensitivity map can ideally show regions that are key to the prediction. 
 
\subsubsection{Counterfactuals.}\label{counterfactual} Counterfactuals were introduced by Wachter et al. \cite{Wachter2017-pi} to provide statements of the differences needed to gain the desirable outcome. This method also works by providing an explanation for the output of the model, but this extra capability makes counterfactuals very unique in realm of XAI methods. 

Jakubowski et al. \cite{jakubowski2022roll} developed a predictive maintenance solution for an industrial cold rolling operation. They utilize a semi-supervised algorithm based on the Physics-Informed Auto-Encoder (PIAE). This architecture was physics-informed by applying a list of equations at the beginning of their input data. The output of the equations was appended to the input data of their AE. Their model proved to be more accurate than a base AE. While PIAE has some interpretable aspects already, they applied counterfactuals as an explainability method to show the important features from their algorithm's decisions.

\subsubsection{Explain Like I'm 5 (ELI5).}\label{eli5} ELI5 is a popular method from Github \cite{TeamHG-Memex} maintained by the user TeamHG-Memex and 15 other contributors. This Python library focuses on explaining the weights of a model which also serves as a method for calculating feature importance. While maintaining original methods, ELI5 also provides other explainability method implementations. 

\subsection{Model-Specific}\label{model_specific}
This section describes the explainable methods in our population that base the explanations on the properties of the architecture it intends to explain. These methods are known as \emph{model-specific}\cite{Belle2021-nd}. Here we discuss methods that take advantage of the architecture for generating explanations such as CAM and GradCAM in Section \ref{cam}, DIFFI in Section \ref{diffi} and more. 

\subsubsection{Class Activation Mapping (CAM) and Gradient-weighted Cam (GradCAM).}\label{cam} CAM was introduced by Zhou et al. \cite{Zhou2016-gv} as a method of global explainability for convolutional neural networks (CNN). The map that is created indicates the image regions that are used by the CNN to identify the target category. CAM does this by utilizing a global average pooling (GAP) layer in the CNN architecture which outputs the spatial average of the feature map of the final layer. The pixels with higher values are associated with the pixels in the image associated with the class label. Additionally, Selvaraju et al. \cite{Selvaraju2020-tq} extend CAM to GradCAM by using the gradient information going into the last convolutional layer to understand the importance of the features. 

GradCAM has been validated through different studies via comparison and metrics. Mey et al. \cite{mey2022explainable} focuses on the plausibility of XAI for explaining a CNN. They investigated GradCAM, LRP and LIME as methods of explaining a CNN for anomaly detection. They found non-distinguishable features highlighted by LRP, and they found unimportant features highlighted by LIME. GradCAM was able to highlight the important features that they labeled prior to CNN training. This could point towards model-specific methods outperforming model-agnostic methods when applicable. 

Solis-Martin et al. \cite{Solis-Martin2023-po} present a comparison on LIME, SHAP, LRP, Image-Specific Class Saliency (Saliency Maps) and GradCAM as applied to predictive maintenance datasets such as CMAPSS and batteries. They identify eight metrics for comparison: identity, separability, stability, selectivity, coherence, completeness, congruence and acumen, an evaluation proposed by the authors. When comparing the different methods as applied to a CNN architecture, GradCAM performed the best in regards to the nine metrics.

Oh et al. \cite{Oh2020explainable} propose a fault detection and diagnosis framework that consists of a 1D-CNN for fault detection, class activation maps for fault diagnosis (explainable method) and VAE for implementing user feedback. The CNN utilizes a GAP layer as the output layer due to its ability to maintain the temporal information. This also allows them to use CAM as an explainable method as opposed to GradCAM. The VAE is utilized with the principle of Garbage-In, Garbage-Out logic to minimize the amount of false positives and negatives that would be presented to the users. To verify their method, they apply it to the Ford Motor dataset which is a vehicle engine dataset that contains an amount of noisy data. They show that their model is accurate even in noisy data, and they show that the VAE increases their accuracy. They also show via CAM that the anomalous data is linearly separable, which is found in the VAE. 


\subsubsection{Depth-based Isolation Forrest Feature Importance (DIFFI).}\label{diffi} DIFFI was introduced by Carletti et al. \cite{Carletti2023-ck} as an explanable method for isolation forests. Isolation forests are an ensemble of isolation trees which learn outliers by isolating them from the inliers. DIFFI relies on two hypotheses to define feature importance where a feature must: induce the isolation of anomalous data points at small depth (i.e., close to the root) and produce a higher imbalance on anomalous data points while being useless on regular points \cite{Carletti2023-ck}. These hypotheses would allow explanations for anomalous data which would allow for explanations of outliers or faulty data. 

Berno et al. \cite{berno2021machine} performed anomaly detection for automated rides at entertainment parks. They introduced the idea of providing extra focus specific features by splitting their data into a multivariate set and many univariate sets based on a prior knowledge. They utilized isloation forest to model the multivariate time series with DIFFI explaining the output. They modeled the univariate time series with a Growing When Required (GWR) neural gas network. The multivariate analysis was used for determining anomalies within most of the variables, and the explanations were used to rank the features causing the anomaly. 

Lorenti et al. \cite{lorenti2022cuad} designed an unsupervised interpretable anomaly detection pipeline known as Continuous Unsupervised Anomaly Detection on Machining Operations (CUAD-MO). CUAD-MO consists of 4 parts: data segmentation and feature extraction, unsupervised feature selection via Forward Selection Component Analysis (FSCA), anomaly detection via Isolation Forest, and post-hoc explainability via DIFFI. Their feature extraction consisted of adding basic statistics and higher order moments of the signals such as Kurtosis. FSCA iteratively selects features to maximize the amount of variance explained. Finally, the Isolation Forest is used to detect outliers which are handled as faulty events. These are explained via DIFFI. They applied their method to 2 years of computer numerical control data resulting in a 67\% precision rate. 

\subsubsection{LionForests.}\label{lionforest} LionForests were introduced by Mollas et al. \cite{Mollas2022-gf} as a local explanation method specifically for random forests. Their method follows these steps: estimating the minimum number of paths for the accurate answer, reducing the paths through association rules, clustering, random selection or distribution-based selection, extracting the feature-ranges, categorical handling of features, composing the interpretation, and visualizing the feature ranges. The outputs of their method are the interpretations in the form of IF-THEN rules and visualizations of the features. 

Mylonas et al. \cite{mylonas2022local} aimed to alleviate the non-explainable nature of random forest by applying an expanded version of LionForests to fault diagnosis. They expanded LionForests into the realm of multi-label classification by applying three different strategies: single label, predicted labelset, and label subsets. Single label aims at explaining every individual prediction (local); predicted labelset aims at explaining all predictions (global); and label subsets aim at explaining based on frequently appearing subsets of predictions. With their expansion, their attention is focused on multiple machine failure datasets, but specifically the AI4I dataset\cite{misc_ai4i_2020_predictive_maintenance_dataset_601}. They utilized accuracy metrics such as precision, and they provided metrics for their explanations such as length of explanations and coverage of data.  One of the more notable elements of their work involves comparing their XAI algorithm to other algorithms, namely global and local surrogates and Anchors.

\subsubsection{Saliency Maps.}\label{salience} Saliency maps were introduced by Simonyan et al. \cite{Simonyan2013-dk} as a method for explaining CNN outputs. Given an input and a model, saliency maps rank the pixels of the input based on their influence on the output of the model. This is done by approximating the output with a linear function in the neighborhood of the input by using the derivative of the scoring function with respect to the input. This approximation is the saliency map.

\subsubsection{Autoencoder-based Anomaly Root Cause Analysis (ARCANA).}\label{arcana} ARCANA was introduced by Roelofs et al. \cite{roelofs2021autoencoder}. They noticed that autoencoders were a popular method of detecting anomalies in their target domain, wind turbines; however by themselves, autoencoders are not interpretable. To overcome this lack of interpretability, they implement ARCANA as a way of explaining the cause of the reconstruction error of an autoencoder. ARCANA works by minimizing a loss function that is based on reconstruction. As opposed to measuring the difference between the output of the autoencoder and the input, they add this bias vector to the input data as to have a \emph{corrected input}.  Moreover, the bias shows "incorrect" features based on the output; therefore, the bias would explain the behavior of the autoencoder by showing which features are making the output anomalous. 

Roelofs et al. \cite{roelofs2021autoencoder} also utilize their method for anomaly detection and root cause analysis for wind turbines. They verify that ARCANA provides the most important feature causing the issues with their wind turbines. This method is done by firstly measuring the features reconstruction error. When performing ARCANA, the feature that shows the most importance is the same feature with the largest error. They then show that even when the feature does not appear in the  reconstruction error, ARCANA is able to find feature importance in sensors that are applicable to known anomalies.

\subsection{Combination of Methods}
This section describes the works that used multiple explainability methods.  Some of these works were utilized to just note the differences between the different explainable methods. Other works compared the methods as to determine the better method. This section reviews the works that combine multiple methods without aiming to declare one method as better than another. 

Utilizing multiple explainable methods can be used in a stacked manner or in a simultaneous manner. The stacked manner involves using explainable methods sequentially. In Jakubowski et al. \cite{jakubowski2021explainable} they created a quasi-autoencoder for explainable anomaly detection. A surrogate model of XGBoost was used as a way of simplifying the original model. They achieved a high $R^2$ score using this XGBoost model while adding explainability via TreeExplainer (SHAP). 

More commonly, a simultaneous utilization of explainable methods appears in the literature where the authors obtain multiple explanations from different methods. Khan et al. \cite{khan2022explainable} found the best architecture for their problem of RUL prediction amongst: random forest, SVM, gradient boosting, elastic net GLM and an MLP regressor. After seeing the MLP regressor to have the best performance, they used LIME and SHAP to explain the output. LIME and SHAP did not have the same explanations, but they had similar explanations. Similarly, Jakubowski et al. \cite{Jakubowski2022-hh} performed an experiment testing five architectures and using SHAP and LIME as explainers. The found that SHAP and LIME had different explanations throughout the different neural architectures suggesting a fidelity concern between architectures. 

Like the prior two, Serradilla et al. \cite{serradilla2020interpreting} performed remaining useful life prediction on a bushings testbed. They tested six different models and determined random forest regressor to be the best. They then utilize two explainability methods (ELI5 and LIME) to show global and local feature importance of driving model development. Brito et al. \cite{brito2022explainable} performed a large experiment that applied many unsupervised learning algorithms for fault detection and fault diagnosis. They showed that Local-DIFFI and SHAP seemed to be mostly in agreement about the explanation for the model's output, but they did not move further in asking which is better. 

Ferraro et al. \cite{Ferraro2023-jv} focused on analyzing the effectiveness of explainability methods on the predictions of a recurrent neural network based model for RUL prediction. Notably, the model performed well, but the focus was on the explainable methods SHAP and LIME. A quantitative analysis was performed using three metrics: identity, stability and separability. This showed: (1) LIME was unable to give identical explanations for identical instances; (2) LIME more than SHAP gave similar explanations to instances in the same class; and (3) LIME and SHAP were able to give different explanations for instances in different classes.

Li et al. \cite{Li2021-ri} aimed at integrating explainability into an AutoML environment used for vehicle data. They tested four different AutoML platforms: AutoSklearn, TPOT, H$_2$O, and AutoKeras. They performed two different experiments where they provided different subsections of their dataset with both resulting in TPOT performing the best in accuracy. Finally, they apply LIME and SHAP to the resulting model to explain a local sample and the whole model. Their work results in a defined workflow for an automatic predictive maintenance system that includes explainability. 

\section{Interpretable ML in Predictive Maintenance}\label{sec:iml_pdm}
\begin{figure*}[!tbp]
  \centering
  \includegraphics[width=\textwidth]{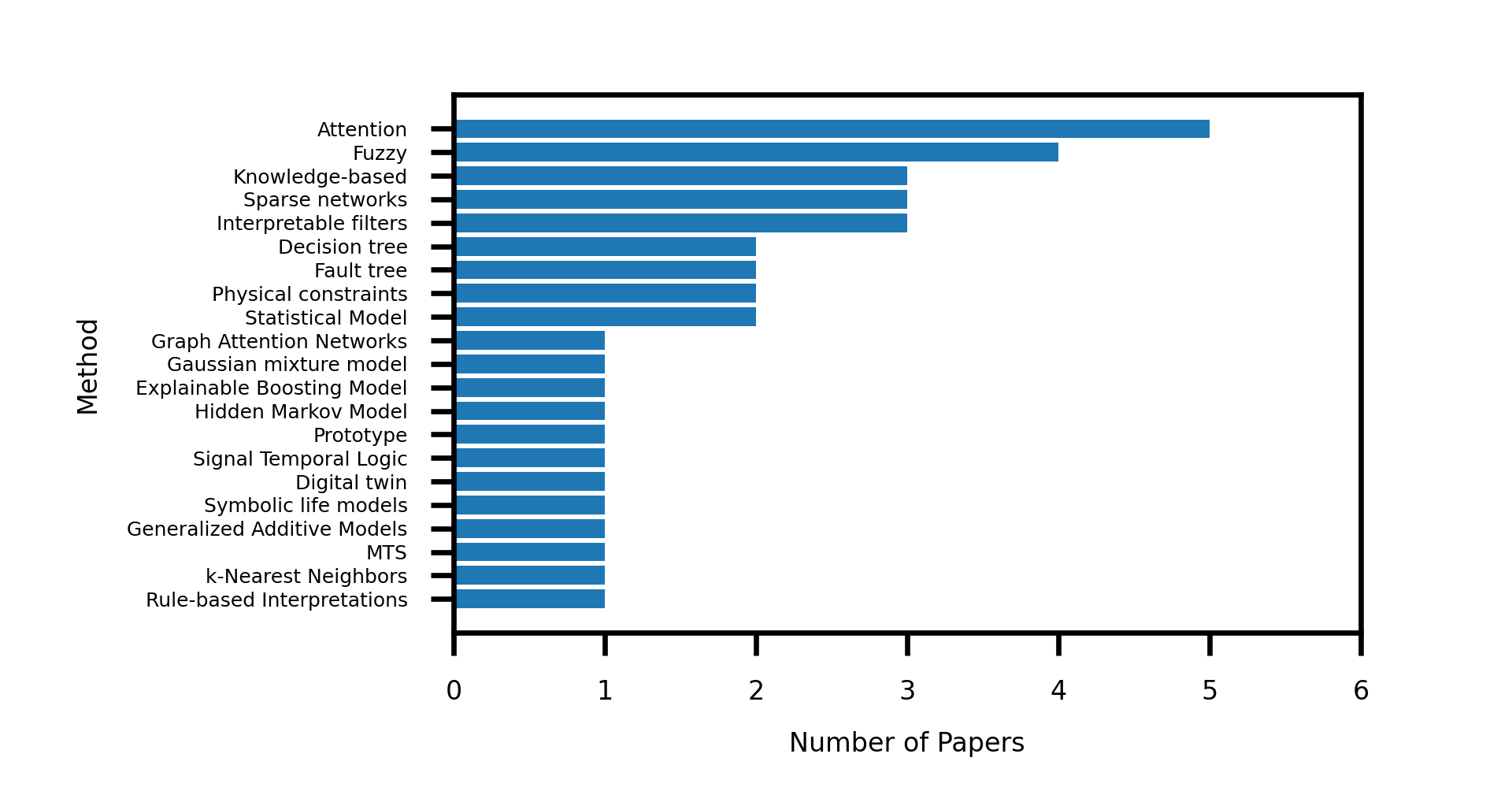}
    \caption{Use of iML Methods}
    \label{fig:imlMethods}
\end{figure*}
Interpretable machine learning (iML) encompasses many methods whose inner-workings are understandable without requiring a post-hoc method for explanation generation. These methods can be interpreted by the target audience without the need of separate methods to serve as a translator between the model and the person. iML methods namely consist of architectures that can have human-readable outputs such as rule-based systems, simple visual representations such as decision trees and simple networks or physical mappings that are intelligible to the user.

\subsection{Attention.}\label{attention} Attention was introduced by Vaswani et al. \cite{Vaswani2017-az} as a method of natural language processing. This attention module gets extended to introduce the transformer architecture that has led to many famous models such as GPT. The weights from the attention modules can be visualized to allow interpretation of the aspects the architecture is focusing. 

Xia et al. \cite{xia2022fault} and Hafeez et al. \cite{hafeez2022dtcencoder} tackled interpretable fault diagnosis in two separate ways. Xia et al. looked at hierarchical attention by grouping the features by systems and subsystems. They utilized BiLSTM encoders with attention to obtain important features where the attention components added interpretability. Hafeez et al. created an architecture known as the DTCEncoder to learn low level representations of multivariate sequences with attention. It utilized the Diagnostic Trouble Codes (DTC) commonly found in predictive maintenance problems as a class label for fault diagnosis. Dense layers were used to translate the encoded latent space from DTCEncoder into a probability distribution for the different DTCs. The latent space was learned using attention mechanisms and could be used to add interpretability of why the network output the DTC. 

For interpretable fault prediction, Wang et al. \cite{wang2021ifp} proposed a two-stage method based on anomaly detection and anomaly accumulation. The anomaly detection module was made using a CT-GAN to train a discriminator on limited data, i.e., faults. The anomaly scores from the CT-GAN were fed into the anomaly accumulation module based on an Attention-LSTM. This modeled the temporal dependencies of the anomaly scores while the attention mechanism was used to give importance to different anomalies at different time steps. Their model outperforms models such as SVM and LSTM on prediction and DTW on classification.

Xu et al. \cite{xu2021deep} was not only interested in anomaly detection, but also anomaly precursor detection, early symptoms of an upcoming anomaly. They argued that detecting precursors is useful for early prediction of anomalies to better understand when and what kind of anomaly will occur. They proposed Multi-instance Contrastive learning approach with Dual Attention (MCDA) to tackle the problem of anomaly precursor detection. MCDA combined multi-instance learning and tensorized LSTM with time-dependent correlation to learn the precursors. Additionally, the dual-attention module produced their interpretable results. This approach had high accuracy results, and their attention mechanism provided variables which are explanatory for the results. Importantly, they verified these explanations with domain experts. 

\subsection{Fuzzy-based.}\label{fuzzy} Fuzzy logic was introduced by Dr. Lofti Zadeh \cite{Zadeh1988-ha} as a way of understanding the approximate mode of reasoning as opposed to the exact. Following this approximate model of understanding, all knowledge would come with a degree of confidence as opposed to a statement being 100\% in a category. This adds some interesting and useful components to machine learning as these in-between categories can be utilized in a way that is different from having all information fall strictly into one category. 

Fuzzy-based methods apply fuzzy logic in different ways. Lughofer et al. \cite{lughofer2022transfer} and Kothamasu et al. \cite{kothamasu2007adaptive} used type 1 fuzzy logic. Lughofer et al. proposed a framework of representation learning based on transfer of fuzzy classifiers. The transfer learning matched the distributions between the source data and the target task using fuzzy rule activation. This was done by feeding the model all of the source data and the healthy data from the target domain. Through this training, the model classified unseen healthy and unhealthy data from the target task. Their model did not outperform all black box models; however, it was in the upper ranks of performance while bringing interpretability to the user.  

Additionally, Kothamasu et al. \cite{kothamasu2007adaptive} presented a Mamdani neuro-fuzzy modeling approach for two use cases, bearing fault detection and aircraft engine fault diagnosis. They chose this model as it has the characteristics of being adaptive, flexible, lucid, and robust. Their model consists of five layers: input, linguistic term input, rules, linguistic terms output, and defuzzification. 
As the rules can become undistinguisable through training, they utilized Kullback-Leibler mean information to refine the rules. 

Fuzzy-based methods can also take the form of higher-order fuzzy logic as seen by Upasane et al. \cite{upasane2021big,Upasane2023-al}. They proposed a type 2 fuzzy logic system for fault prediction to allow interpretability \cite{upasane2021big}. Additionally, the Big-Bang Big-Crunch (BB-BC) evolutionary algorithm was used for optimizing the number of antecedents of their fuzzy logic system. This was optimized for minimizing the RMSE of their system. Their system was able to get a very low RMSE with 100 rules and six antecedents per rule. 

Upasane et al. \cite{Upasane2023-al} extended their previous work \cite{upasane2021big} to include most of the faults that can occur as well as proposing an explainable framework. While maintaining accuracy with more faults is noteworthy, the experiment's measurement of users' trust was quite unique compared to the literature. They observed that 80\% of the respondents agreed or strongly agreed with having trust in the interpretable system. This trust is attributed to the explainable framework and interpretable nature of their architecture; moreover, the interface is noted to provide helpful insights to the users that would minimize downtime of the assets. 

\subsection{Knowledge-based.}\label{knowledge} In this paper, \emph{knowledge-based} approaches include methods such as knowledge-graphs, knowledge-based systems, knowledge graphs, etc. Knowledge-based approaches focus on a symbolic representation of the data that one can find in a source of data. These representations consist of connections between different features where the links take the form of a link when discussing graphs or production rules when discussing production systems. These methods produce interpretation by providing these connections within the features, usually in the form of natural language. 

Xia et al. \cite{Xia2023-bg} proposed a maintenance-oriented knowledge graph to apply for predictive maintenance of oil drilling equipment. Once they had the maintenance-oriented knowledge graph, an attention-based compressed relational graph convolutional network (ACRGCN) was used to predict solutions for different faults by predicting links between knowledge. This method also explained faults due to its knowledge-graph that maps different symptoms and maintenance requirements. Even though knowledge-graphs have inherent interpretability, they created a question-answer system that allowed the user to query the graph. 

Salido et al. \cite{salido2004comparison} created a fuzzy diagnosis system based on knowledge-based networks (KBN) and genetic algorithms (GA). The KBN constructed fuzzy rules using neural learning where the input is the features and the following layers are OR neurons and AND neurons. To determine the optimal number of neurons, they used a GA. Importantly in their GA, they added a metric to measure simplicity of their rules by making more concise rules. With their architecture, they could 1) detect a fault and 2) explain the fault using an IF-THEN rule which can be used as a method of root cause analysis. 

Cao et al. \cite{cao2020using} created an approach based on knowledge-based systems for anomaly prediction. Their method is broken into three parts: pruning of chronicle rule base, integration of expert rules, and predictive maintenance. Pruning of chronicle rule base consists of mining the rules with frequent chronicle mining, translating the rules into SWRL rules, and using accuracy (how many true rules) and coverage (how many true encompassing rules) to select the best quality rules. The integration of expert rules involved receiving input from the experts and placing the same restrictions on their rules. Finally, the rules were used for anomaly prediction of semiconductors.

\begin{table}[ht]
\centering
\caption{Interpretable Methods from the Literature}
\begin{tabularx}{0.5\textwidth}{| d | a |}
\hline
{\textbf{Method}}&{\textbf{Articles}}\\
\hline
Attention (\ref{attention}) & \cite{Jing2022-yq,xia2022fault,hafeez2022dtcencoder,wang2021ifp,xu2021deep} \\
Fuzzy (\ref{fuzzy}) & \cite{lughofer2022transfer,kothamasu2007adaptive,Upasane2023-al,upasane2021big} \\
Knowledge-based (\ref{knowledge}) & \cite{Xia2023-bg,salido2004comparison,cao2020using} \\
Sparse Networks (\ref{sparse}) & \cite{pu2023restricted, Beretta2021-gk, langone2020interpretable} \\
Interpretable Filters (\ref{interpretable_filter}) & \cite{li2022waveletkernelnet,li2023wavcapsnet,ben2022end} \\
Decision Tree (\ref{decision_trees}) & \cite{Amram2021-gt,panda2023ml} \\
Fault Tree (\ref{fault_trees}) & \cite{verkuil2022automated,Waghen2022-na} \\
Physical Constraints (\ref{physical}) & \cite{Tod2023-cf,xu2022physics} \\
Statistical Model (\ref{statistical_models}) & \cite{wang2020online,yao2023integrated} \\
Graph Attention Networks (\ref{GAT}) & \cite{liu2023causal} \\
Gaussian Mixture Model (\ref{GMM}) & \cite{csalodi2021mixture} \\
Explainable Boosting Machine (\ref{EBM}) & \cite{Jakubowski2022-hh} \\
Hidden Markov Model (\ref{HMM}) & \cite{Abbas2022-mm} \\
Prototype (\ref{prototype}) & \cite{ming2019interpretable} \\
Signal Temporal Logic (\ref{stl}) & \cite{chen2020temporal} \\
Digital Twin (\ref{digital_twin}) & \cite{Mahmoodian2022-zk} \\
Symbolic Life Model (\ref{slm}) & \cite{Ding2021-ez} \\
Generalized Additive Model (\ref{gam}) & \cite{yang2022noise} \\
MTS (\ref{mts}) & \cite{scott2023computational} \\
k-Nearest Neighbors (\ref{knn}) & \cite{konovalenko2022generating} \\
Rule-based Interpretations (\ref{rule_based_interpret}) & \cite{Dhaou2021-qx} \\
\hline
\end{tabularx}
\label{tab:interpretable}
\end{table}

\subsection{Interpretable Filters.}\label{interpretable_filter} Interpretable filters are a concept that brings specific waveforms to a CNN architecture as a way of showing what signals are being learned. As explained in Ravanelli and Bengio \cite{ravanelli2018speaker}, the first layer of a CNN appears to be important for waveform-based CNNs. In using these interpretable filters that take the form of common waveforms, one can begin to understand the behavior of the CNN if one understands the behavior of the waveform. 

Li et al. \cite{li2022waveletkernelnet} aimed to improve CNN-based methods for PHM by addressing the black box problem. They proposed the Continuous Wavelet Convolution (CWC) layer which is designed to make the first layer of a CNN interpretable. It does this by using a library of filters that have physical meanings which are convolved on the input signal. These convolutions can be traversed along the series and projected into a two-dimensional time and scale dimension. Its performance was compared with a CNN with different wavelets, and their findings were two-fold. Firstly, the performance of the CNN with a CWC layer showed better performance than a CNN without. Lastly, the CWC learned a well-defined waveform while the one without learned what looked to be a noisy and uninterpretable representation.

Li et al. \cite{li2022waveletkernelnet} built on their previous work by examining compound faults. They designed an interpretable framework called wavelet capsule network (WavCapsNet) which utilizes backward tracking. This network has 1) interpretable meaning from the wavelet kernel convolutional layer, 2) capsule layers that allow decoupling of the compound fault, and 3) backward tracking which helps interpret output by focusing on the relationships between the features and health conditions. Not only was their framework able to achieve high accuracy on all conditions, including compound faults, but also they showed that the backward tracking method can decouple the capsule layers effectively.

Ben et al. \cite{ben2022end} proposed a new architecture, SincNet, that trains directly on the raw vibration signals to diagnose bearing faults. Their architecture utilized interpretable digital filters for CNN architectures. They reduced the number of trainable parameters and extracted meaningful representations by having the predefined functions serve as the convolution. When comparing the performance to a CNN, the SincNet had a higher precision and reached convergence faster.

\subsection{Decision Trees.}\label{decision_trees} Decision trees encompass both classification and regression trees that date back to the first regression tree algorithm proposed by Morgan and Songquist \cite{morgan1963problems}. Decision trees create a tree-based architecture where each set of children of each node is split using a feature. To produce an output, a decision tree algorithm starts at the root of the tree and proceeds down the tree by evaluating the feature that is used for splitting. The output corresponds to the final leaf node that the decision trees reaches on its path. 

Amram et al. \cite{Amram2021-gt} utilized two types of decision trees, optimal classification trees \cite{bertsimas2017optimal} and optimal survival trees \cite{bertsimas2022optimal}. Their goals included predicting the RUL of long-term health of hard drives, predicting RUL of the short-term health of hard drives, predicting failure classification in short-term health of the hard drives and performing similar experiments with limited information. Their results showed that they could gather better results using separate models for the tasks as opposed to using one model. They also showed the interpretable methods shared many of the important features for the different tasks. 

Panda et al. \cite{panda2023ml} aimed at tackling the problem of commercial vehicle predictive maintenance by designing an interpretable ML framework. To simplify their problem, they solely looked at the air compressor system. By looking at the air compressor system, they ran a broad experiment that analyzed different configurations of models and data. The C5.0 with boosting model performed the best, and the inclusion of Diagnostic Trouble Codes with the sensor data raised the performance metrics.

Simmons et al. \cite{Simmons2021-hj} argued that the dynamics of a time-series are in themselves discriminative of health or failure. Additionally, the dynamics are interpretable because they are derived directly from the information. These ideas were translated into the data mining domain by creating features that represent shorter time series in the temporal, spatial, and mixed domains. The features went through a rank-based selection process which gathered features that were statistically different between classes. These features were used to train a Light Gradient Boosting Machine (LightGBM) which is a type of gradient boosting decision tree introduced by Ke et al. \cite{ke2017lightgbm}. This method allows for constant monitoring of feature importance during training which can be used for interpreting the results. 

\begin{table*}[ht!]
\centering
\caption{Examples of Articles from Sample Population}
\begin{tabularx}{\textwidth}{| X | X | X |}
\hline
{\textbf{Title}}&{\textbf{Objective}}&{\textbf{Contribution}}\\
\hline
Impact of Interdependencies: Multi-Component System Perspective Toward Predictive Maintenance Based on Machine Learning and XAI \cite{gashi2023impact} & Perform predictive maintenance by modeling interdependencies and test their importance & Showed with statistical significance that interdependency modeling increases performance and understandability of a model \\ \hline

Explainable and Interpretable AI-Assisted Remaining Useful Life Estimation for Aeroengines \cite{Protopapadakis2022-ye} & Compute RUL of the CMAPSS turbofan dataset with LIME explaining the performance & Showed that LIME performed poorly when applied to segments with no faults but performed well when labeling features with failing sequences \\ \hline

Explainability-driven Model Improvement for SOH Estimation of Lithium-ion Battery \cite{Wang2023-px} & Perform predictive maintenance by embedding explanations into the training loop & Introduced the idea of explainability-driven training for predictive maintenance \\ \hline

Online Anomaly Explanation: A Case Study on Predictive Maintenance \cite{ribeiro2022online} & Apply XAI methods to the online learning process & Showed that local and global explanations could be added into the online learning paradigm \\ \hline

Explaining a Random Forest with the Difference of Two ARIMA Models in an Industrial Fault Detection Scenario \cite{glock2021explaining} & Utilize two ARIMA surrogate models to explain the capabilities of a random forest model & Introduced a method of sandwiching a model between two surrogates to show where a model fails to perform well \\ \hline

Edge Intelligence-based Proposal for Onboard Catenary Stagger Amplitude Diagnosis \cite{ERRANDONEA2023103781} & Test XAI on edge computing for fault diagnosis & Provided a method of performing XAI in an edge computing example coupled with AutoML libraries\\ \hline

Explainable AI Algorithms for Vibration Data-based Fault Detection: Use Case-adapted Methods and Critical
Evaluation \cite{mey2022explainable} & Discover the plausibility of XAI methods explaining the output of CNN architectures & LRP showed non-distinguishable features, LIME showed unimportant features, and GradCAM showed the important features \\ \hline

On the Soundness of XAI in Prognostics and Health Management (PHM) \cite{Solis-Martin2023-po} & Compare different XAI methods for the CMAPSS and lithium-ion battery dataset & Showed different metrics for comparing explanations generated by different XAI methods and showed GradCAM to perform the best on CNN architectures \\ \hline

Interpreting Remaining Useful Life Estimations Combining Explainable Artificial Intelligence and Domain Knowledge in Industrial Machinery\cite{serradilla2020interpreting} & Perform RUL of bushings through multiple different models and explanatory methods & Showed the importance of applying global and local explanations to interpret performances of models from all aspects \\ \hline

Evaluating Explainable Artificial Intelligence Tools for Hard Disk Drive Predictive Maintenance \cite{Ferraro2023-jv} & Analyze the effectiveness of explainability methods for recurrent neural network based models for RUL prediction & Utilized three metrics to compare explanations from LIME and SHAP and showed where each of them shine over the others \\ \hline

Automatic and Interpretable Predictive Maintenance System \cite{Li2021-ri} & Aimed to integrate explainability into an AutoML environment & Defined a workflow for an automatic explainable predictive maintenance system \\ \hline

DTCEncoder: A Swiss Army Knife Architecture for DTC Exploration, Prediction, Search and Model interpretation \cite{hafeez2022dtcencoder} & Perform fault detection by classifying DTCs  & Designed the DTCEncoder that utilizes an attention mechanism to provide an interpretable latent space as to why the a DTC is output \\ \hline

Deep Multi-Instance Contrastive Learning with Dual Attention for Anomaly Precursor Detection \cite{xu2021deep} & Perform anomaly detection and anomaly precursor detection & Performed anomaly precursor detection through multi-instance learning with verified explanations through domain experts \\ \hline

A Type-2 Fuzzy Based Explainable AI System for Predictive Maintenance Within the Water Pumping Industry \cite{Upasane2023-al} & Utilize an evolutionary algorithm to optimize their fuzzy logic system for fault prediction & Used a type 2 fuzzy logic system and evolutionary optimization to generate fuzzy rules for fault prediction \\ \hline

Waveletkernelnet: An Interpretable Deep Neural Network for Industrial Intelligent Diagnosis \cite{li2022waveletkernelnet} & Improve CNN-based methods for PHM & Designed the Continuous Wavelet Convolution to add physical interpretations to the first layer of CNN architectures \\ \hline

Restricted Sparse Networks for Rolling Bearing Fault Diagnosis, \cite{pu2023restricted} & Perform fault detection using a sparse network & Explored the Restricted-Sparse Frequency Domain Space and used the transform into this space to train a two-layer network that performs equal to a CNN-LSTM \\ \hline

Interpretable and Steerable Sequence Learning via Prototypes \cite{ming2019interpretable} & Construct a deep learning model with built-in interpretability for fault diagnosis via DTCs & Introduced Prototype Sequence Network (ProSeNet) which uses prototype similarity in the training of the network and justified the interpretability of their approach via a user study on Amazon MTurk \\ \hline

Causal and Interpretable Rules for Time Series Analysis \cite{Dhaou2021-qx} & Perform predictive maintenance while utilizing causal rules for explanations & Designed Case-crossover APriori algorithm for predictive maintenance which showed both higher performance occurs when having rules that are additive and subtractive to an output\\

\hline
\end{tabularx}
\label{tab:articles}
\end{table*}

\subsection{Fault Trees.}\label{fault_trees} Fault trees were introduced by H.A. Watson at Bell Labs in 1961 \cite{watson1961launch}. Fault trees were introduced as an understandable model that can learn complex systems and perform root cause analysis. They are tree-like structures that are created using different types of nodes: basic events, gate events, condition events, and transfer events. Basic events are the nodes that represent either a failure event or a normal operating event. Gate events are the logic combining nodes and consists of AND, OR, Inhibit, Priority and Exclusive OR. Condition events represent conditions that must occur for a gate event to occur. Transfer events are nodes that point to somewhere else in the tree. With all of these gates, fault trees are able to learn root causes for different faults that can occur in a system. 

Verkuil et al. \cite{verkuil2022automated} noticed that fault trees are made via human intervention. With the idea of automating the process, they applied the C4.5 tree combined with LIFT to create fault trees for domestic heaters. C4.5 is used to learn the failure thresholds of the sensor data. LIFT creates fault trees in an iterative process using the learned features. While they do not provide a performance metric, they note that their method cannot be optimal for the reasons of oversimplifying the problem and using a greedy heuristic. However, domain experts weighed in on the explanations provided in a positive manner.

Waghen et al. \cite{Waghen2022-na} utilized fault trees to perform interpretable time causality analysis. Their methodology consisted of building multiple logic trees for each subset of data. These logic trees were aggregated into one fault tree representing the multiple trees. They performed interpretatable time cause analysis by going through each variable in the fault tree. By traversing the fault tree, they were able to extrapolate rules that can model the causality through time towards faults. 

\subsection{Physical Constraints.}\label{physical} Physical constraints are used to bring real-life limitations to the data-driven models. This can be in the form of mapping the input and output of the architectures to physical components, or more commonly, utilizing known physics information or equations about the real-life system in the architecture of their model in some way. 

The methods of applying physical constraints can be seen in different forms, namely model-based approaches and physics-informed approaches, which need to be differentiated. Model-based approaches are created to model a system without the training of a network with the data provided, separate from data-based models \cite{Tod2023-cf}. These model-based approaches have physical constraints as they have to model the mathematical properties of the system. Physics-informed models aim to combine model-based and data-driven approaches by attaching the mathematical properties of the system to the data in data-driven approaches \cite{xu2022physics}. 

Tod et al. \cite{Tod2023-cf} implemented a first-principle model-based approach to assess the health of solenoid operated valves. Compared to other first-principle models, their improved model takes other degradation effects into account, namely shading ring degradation and mechanical wear. The method extracts three condition indicators which allows them to detect problematic signals that can be directly mapped to physical components through their model. 

Wang et al. \cite{wang2020online} performed fault diagnostics of wind turbines. Their method was an online method that detected issues with bearings. Coupled with equations that represent the physical aspects of the bearings, they detected issues surrounding clearance of the bearings with high interpretability. Their interpretation specifically showed the different frequencies around the physical features of the bearings. 

Xu et al. \cite{xu2022physics} propose the physics-constraint variational neural network (PCVNN) as applied to external gear pumps. The PCVNN is physics-informed asymmetric autoencoder where the encoder is a stacked CNN, BiLSTM, Attention network while the decoder is a generative physical model. This would allow for an NN to learn the data, and it would allow the physical model to represent the learned patterns in a way that is consistent with the physics of the problem.

\subsection{Statistical Methods.}\label{statistical_models} Statistical methods are using for explaining by analyzing different features along different classes using statistical tests, such as Student's t-test \cite{student1908probable}, Pearson's chi-squared test \cite{pearson1900x}, etc.

Yao et al. \cite{yao2023integrated} proposed a framework with interpretable and automatic approaches that consisted of solely statistical processing. Their method proposed kurtosis-energy metric to define key sub-bands, a new health index of these sub-bands, a joint statistical alarm and fault identification strategy. Additionally, they proposed a health phase segmentation strategy for health phase assessment and degradation pattern analysis. This method involved analyzing the data on the time-frequency domain and suppressing the disturbing components such as noise. This analysis was able to help form the sub-bands for monitoring the current state. If it fails statistical tests, then an anomaly is detected. They tested their method on the PHM 2012 rolling bearing dataset, and they reported very low false positives.

\subsection{Graph Attention Networks (GATs).}\label{GAT} GATs were introduced by Velivckovic et al. \cite{velivckovic2017graph} as a way of combining self-attention layers with graph-structured data. This is done by applying attention layers where nodes can attend whole neighborhoods of previous graph nodes. While this comes with many benefits, the main two come from the benefits that other architectures gain from attention mechanisms and the retraction of needing prior knowledge of the graph structure. 

Liu et al. \cite{liu2023causal} designed a framework for fault detection based around the Graph Convolutional Network and Graph Attention Networks. They propose the Causal-GAT. Causal-GAT is comprised of two parts: causal graph construction and DC-Attention for extracting features and detection. The causal graph construction uses causal discovery methods and/or prior expertise to encode monitoring variables into a directed acyclic graph. The Disentangled Causal Attention (DC-Attention) aggregates the causal variables for generating representations of the effect variables. The DC-Attention outputs the system status (faulty or not faulty). They then utilize a custom loss function that calculates the distance between the current support of representations and its theoretically disentangled support. 

\subsection{Gaussian Mixture Model.}\label{GMM} As described by Reynolds \cite{reynolds2009gaussian}, Gaussian mixture models (GMMs) is a probability density function designed as a weighted sum of Gaussian component densities. The component densities are created using the mean vector and covariance matrix of the data while the mixture weights are estimated. GMMs are commonly used due to their capability of representing information via a discrete set of Gaussian functions to improve modeling of larger distributions. These models can be labeled as interpretable as the models directly represent the distributions of the features. These models can then be directly used to explain the features. 

Csalodi et al. \cite{csalodi2021mixture} performed survival analysis via a Weibull distribution by representing the operation signals as a Gaussian mixture models and the parameters of the Weibull model via clustering. Specifically, their method used an expectation-maximization algorithm which consists of two parts. The expectation step determined the probability that a data point belongs to any cluster given the survival time and parameters while assuming the clustering is correct. The maximization step updated the parameters for the Gaussian mixture models and the Weibell distribution to better represent the data. When applying their method to lithium-ion batteries, they represented distributions of unhealthy batteries quite accurately while healthy batteries were less well-represented. This occurred due to the large category of healthy data which was harder to represent in one small model while the unhealthy data could be easily represented when isolated. 

\subsection{Explainable Boosting Machine (EBM).} \label{EBM}EBM were introduced by Nori et al.  \cite{nori2019interpretml} as a \emph{glassbox model}, another term for interpretable model, with similar accuracy to that of state-of-the-art blackbox algorithms. EBM is a type of generalized additive model that learns each feature's function using techniques such as bagging. Additionally, it can detect interactions between features and include those pairs of terms by learning functions of combinations of features. Because of its nature as an additive model, the features can be explained by their impact on the outcome. 

\subsection{Hidden Markov Model.}\label{HMM} HMMs were introduced by Baum and Petrie \cite{baum1966statistical} and can be described as a statistical state-space algorithm \cite{ramezani2021hmm}. HMMs represent the learning as a statistical process that transitions between states, and HMMs represent the output as separate states that extend from the transitional states. HMMs, as a statistical process, can discern hidden states from the data that may not be readily apparent. They are also capable of learning combinations of sensor data, leveraging confounding variables, and executing dimensionality reduction to simplify the complexity of the data. \cite{Abbas2022-mm}. 

Abbas et al. \cite{Abbas2022-mm} combined the input-output HMM with reinforcement learning to make interpretable maintenance decisions. Their hierarchical method consisted of two steps. The input-output HMM filters the data and detects failure states. Once the failure state was detected, the deep reinforcement learning agent learned a policy for maintenance based on the failures. The first challenge of this approach involves representing predictive maintenance as a reinforcement learning problem. This is done by representing the potential actions as hold, repair, or replace, creating a reward function based on holding, early replacement and replacement after failure, and measuring the cost based on these reward functions. The HMM is used for interpreting the output of their model by observing the features that led the model into detecting a failure state. 

\subsection{Sparse Networks.}\label{sparse} Sparse networks are neural networks that are limited in their architecture. Large deep neural networks are inherently blackbox models; however, interpretable whitebox models can take the form of very simple neural network models such as linear regression or logistic regression models. As the models are simple, the impacts of the input features can be seen as they are propagated through the network. 

Beretta et al. \cite{Beretta2021-gk} utilized two different models for predictive maintenance: a gradient-boosting regressor to model the normal data and an isolation forest to model the fault data. The output of these are merged with a mean average of the temperature readings to create a score of failure. The authors praise the simplicity of the algorithms as the source of interpretability in their method. 

Pu et al. \cite{pu2023restricted} explored a new frequency domain space they call the restricted sparse frequency domain space (RSFDS) for rolling bearing faults. The RSFDS breaks down the features into a space that is made of real and imaginary points. This space is able to visualize boundaries that have physical meanings to the faults. They use a simple two-layer neural network to these points, and they achieve high performance equal to that of a CNN-LSTM with less memory and CPU usage. 

Langone et al. \cite{langone2020interpretable} proposed a model for interpretable anomaly prediction based on a logistic regression model with elastic net regularization. Their method is made of 3 steps: data preparation, learning and refinement of the prediction model. In the data preparation phase, they categorize the data using included statistics, apply windowing to the data, and finally mark the windows as either being anomalous or not. The learning phase consists of learning the relevant features from the windowed data. This includes considering the feature distributions across failures and non-failures and measuring the distance according to the Kolmogorov-Smirnov metric. The refinement of prediction model phase consists of the training and utilization of the logistic regression model. Coupled with elastic net regularization, this model selected a smaller subset of the original data and captures the variable correlations. They applied their method to a plunger pump in a chemical plant and produced relative good and consistent scores. 

\subsection{Prototype Learning.}\label{prototype} Prototype learning, as described by Ming et al. \cite{ming2019interpretable}, is a form of case-based reasoning that determines the output of an input by comparison to a representative example. Determining the best prototypes is a problem itself, but the interpretability it bring is apparent. The output of a specified input would be similar to its most similar prototype's output; therefore, the reason that the input data has a certain output is due to the output of a very similar piece of data. This brings interpretability via comparison to the prototype. 

Ming et al. \cite{ming2019interpretable} used the concept of prototype learning to construct a deep learning model with built-in interpretability. They introduced the prototype sequence network (ProSeNet) for a multi-class classification problem of fault diagnosis via diagnostic trouble codes. The model consists of a sequence encoder that is based on a recurrent architecture. The hidden state is fed into a prototype layer that determines how similar the hidden state is to prototypes in the form of a similarity vector. The network then outputs a prediction probability for the different classes based on the similarity vector. Interpretability can be conceived via the prototypes that are most similar to the input. They justified the interpretability of their model by using Amazon MTurk and surveying the users about the interpretability. They also studied how the input of human knowledge would affect the interpretability. They showed that including the human feedback improved the interpretability of their network in a post-study of different Amazon MTurk users. 

\subsection{Signal Temporal Logic (STL).}\label{stl} Introduced by Maler and Nichovic \cite{maler2004monitoring}, STL as a type of temporal logic that is used for \emph{dense-time real-valued signals}. STL is defined as predicates over atomic propositions. These STL rules are formed by applying Boolean filters for these atomic propositions that transforms a signal into a Boolean signal. This involves considering: the filter that is being applied, the length of the signal, the sampling of the signal and any additional desired samples. We refer the reader to Maler and Nichovic \cite{maler2004monitoring} Section 4 for an example. 

Chen et al. \cite{chen2020temporal} performed fault diagnosis on a furnace using internet-of-things, reinforcement learning, and signal temporal logic. Their algorithm takes in the STL grammar and labeled input data, and it outputs an optimal STL formula. The agent chooses a formula from the agenda and adds it to a chart based on the current policy. The evaluator evaluates the performance of the formula on the input. The learner updates the policy function according to the performance. The agenda is updated based on the formulas in the chart. They utilize an MDP to construct the agenda-based formulas while the reinforcement learning solves the problem. They apply their method to multiple faults demonstrating good robustness results, fast runtimes, and statistically significant performances.

\subsection{Digital Twin.}\label{digital_twin} Digital twins originated in 2002 as described by Grieves and Vickers \cite{grieves2017digital} as a way of creating a digital construct that describes a physical system. Moreover, digital twins consist of two systems: a physical system that is represented by the asset and a digital system that holds the information about the physical system. Using digital twins, one can observe the performance of the physical system without having the physically observe the asset. 

Mahmoodian et al. \cite{Mahmoodian2022-zk} proposed the use of a digital twin to monitor the infrastructure of a conveyor. Their digital twin consists of taking in real data from different sensors and simulating the data. This data is compared to the real time data to ensure the data is consistent. Their digital twin can display the different information as well as receive input from the users to rate the explanations given. If it is seen as not valid, the digital twin can run simulations surrounding that data to increase its accuracy. 
 
\subsection{Symbolic Life Model.}\label{slm} Symbolic life models aim to alleviate the black box effect by modeling the process learned by mapping relationships and results. Symbolic life models are a form of symbolic regression based on genetic programming. This method creates a tree representation of an equation where the nodes are an input, a mathematical expression or a number. The output of the tree given an input is found by traversing the tree and performing the mathematical expressions as nodes are expanded. The genetic algorithm is used to perform crossovers and mutations based on the different mathematical functions and numbers where the goal is to maximize the tree's performance on a given dataset. For more detailed information, we recommend Augusto and Barbosa \cite{augusto2000symbolic}. 

Ding et al. \cite{Ding2021-ez} proposed the use of symbolic life models, specifically dynamic structure-adaptive symbolic approach (DSASA), as a way of modeling RUL. DSASA combines the evolving methods of symbolic life models with the structure of adaption methods. An initial symbolic life model is created from a genetic programming algorithm and run-to-failure data. This is followed by the dynamic adjustment to the life models based on the performance on real-time information. This creates groups of improved models that can all be used for prediction. The life models are interpretable as they are simple models that perform based on the physical constraints. 

\subsection{Generalized Additive Model (GAM).}\label{gam} Introduced by Hastie and Tibshirani \cite{Hastie1986-zf}, GAMs are a way of estimating a function by summing a list of nonlinear functions in an iterative manner as to become better with accurate local models as opposed to an overarching global model. These local models are smoothed using a series of smoothing functions. Additionally, these local models are independent of one another as they are trained using single features. These local models allow for interpretability as well as importance related to their impacts on the outcome of the GAM. 

Yang et al. \cite{yang2022noise} introduced the Noise-Aware Sparse Gaussian Process as a way of solving the scalability and noise sensitivity issues of normal Guassian Processes. Based on their NASGP algorithm, they developed an interpretable GAM that uses additive kernels and individual features. They applied their method to the IEEE PHM 2012 data challenge in forms of RUL prediction and fault diagnosis. Their method performed well in comparison to other methods and allowed a level of interpretability.  
 
\subsection{Mahalanobis-Taguchi System (MTS).}\label{mts} MTS was introduced by Taguchi and Jugulum \cite{taguchi2002mahalanobis} as a diagnosis and forecasting method. This method bases its discriminative power on the Mahalanobis distance calculation; this method cannot feasibly work if the classes cannot be distinguished this way. The feature space is reduced via orthogonal arrays and signal-to-noise ratios. The orthogonal array contains different subsets of the features. The signal-to-noise ratio measures the abnormality of the feature. Finally, the Mahalanobis distance is maximized by only including the features whose signal-to-noise ratio increases the distance. This maximized distance can be seen as the reason for a diagnosis, which is determined by the features that are used to calculate the distance. 

Scott et al. \cite{scott2023computational} introduced use of the Mahalanobis-Taguchi system for fault detection. MTS utilizes Mahalanobis distance, orthogonal arrays, and signal-to-noise ratios for multivariate diagnosis and prediction. The Mahalanobis space represents the stable operations andyields the difference of an observation from stable. The orthogonal arrays and signal-to-noise ratio is used to diagnose or identify variables responsible for the fault. This method was able to detect roughly 75\% of the faults tested. 

\subsection{k-Nearest Neighbors (kNN).}\label{knn} Originally introduced in 1951 by Fix and Hodges \cite{fix1989discriminatory}, kNN is a supervised learning algorithm that is based on grouping input data with the k most similar other pieces of input data. It represents the input data as a large feature space. The output of some input data is represented by its place in the feature space in relation to the k closest other data points. Small k values lead to less consideration for the output value of the input data; however, it also leads to a more specific output. Larger k values lead to considering more values when determining the output; however, too large k values will make the output less meaningful. 

Konovalenko et al. \cite{konovalenko2022generating} used a modified kNN algorithm for generating decision support of temperature alarms. They tackled three problems associated with kNN: (1) the difficulty associated with sparse regions; (2) the blindness to class boundaries leading to misclassifications; and (3) sensitivity to class overlap. These problems were addressed by adding principles of local similarity and neighborhood homogeneity. Local similarity refers to the idea that a new data is closer to training samples with the same class label. Neighborhood homogeneity is the idea that new data falls into a neighborhood where the class label represents the majority. This method is interpretable through its ability to separate classes of data on a small dimensional graph. 

\subsection{Rule-based Interpretations.}\label{rule_based_interpret} Similar to rule-based explainers presented in \ref{model-agnostic}, rule-based interpretations involve utilizing rules that are learned from the data. Unlike the rule-based explainers, rule-based interpretations remove the black-box from the problem. This allows the rules to be directly learn from the information as opposed to learning from the black-box model and the data.

Dhaou et al. \cite{Dhaou2021-qx} proposed a novel approach that combines case-crossover research design with Apriori data mining. This combination resulted in the Case-crossover APriori (CAP) algorithm for association and causal rules explanation. The case-crossover design describes the way of setting up the problem. They ignored the group of data where nothing goes wrong, and they focused on the subjects that have the class change. In the case of predictive maintenance, a class change would be from healthy to failure data. The case-crossover design looks at the period prior to class change as the control group, and it looks at moments before the class change as the case period. These data points are combined with Association Rule Mining APriori to extract causal rules. These causal rules can be both additive (predictive of truth) and subtractive (predictive of falsehood). Their results show that both additive and subtractive rules help with performance, and they show their algorithm to outperform random forest on the same problem. 

\section{Challenges and Research Directions of Explainable Predictive Maintenance}
\label{sec:challenges}
XAI and iML have been successfully utilized in predictive maintenance on many accounts. Researchers have shown that these methods can add to a prediction in a way that can be used for root cause analysis, validation of faults, etc. The main focus of much of the research focuses on adding explainability to a complex and unexplainable problem. While an important aspect of this field of study, there are multiple facets to the problem that generally go under-represented. 

\subsection{Purpose of the Explanations}
All explanations serve one overarching purpose: produce reasons that make the model's functioning understandable. This information transfer has taken form in visualizations of data distributions, visualizations of feature importance graphs, predictive rules, etc; however, the information is not specific to a target audience. To echo Neupane et al. \cite{Neupane2022-ly}, \emph{"explanations are not being designed around stakeholders".} Not only are the explanations not being designed for stakeholders, but also many explanations do not have a target audience outside of the implicit audience of the model's designer. 

Barredo Arrieta et al.\cite{Barredo_Arrieta2020-xv} provides a list of potential audiences XAI can target. While they go into more detail, some potential target audiences, especially for predictive maintenance, could be the data scientists and developers creating the predictive system, the project managers and stakeholders in the project, or even the mechanics working on the physical systems. These different people may need different types of explanations ranging from more explanations relating to the physical and time domains to higher level abstract information.  

\begin{table}[ht]
\centering
\caption{Explanation Evaluation Metrics from \cite{Miller2019-xc, coroama2022evaluation, sisk2022analyzing, kadirassessing, hoffman2018metrics}}
\begin{tabularx}{0.5\textwidth}{| s | s | X |}
\hline
{\textbf{Metrics}}&{\textbf{Viewpoint}}&{\textbf{Description}}\\
\hline
D & Objective & Difference between the model's performance and the explanation's performance \\
R & Objective & Number of rules in explanations \\
F & Objective & Number of features in explanation \\
S & Objective & The stability of the explanation \\ 
Sensitivity & Objective & Measure the degree in which explanations are affected by small changes to the test points\\
Robustness & Objective & Similar inputs should have similar explanations \\
Monotonicity & Objective & Feature attributions should be monotonic; otherwise, the correct importance is not calculated \\
Explanation correctness & Objective & Sensitivity and Fidelity \\ 
Fidelity & Objective & Explanations correctly describe the model; features and their attribution are correlated \\
Generalizability & Objective & How much one explanation informs about others \\
Trust & Subjective & Measured through user questionnaires \\
Effectiveness & Subjective & Measures the usefulness of the explanations \\ 
Satisfaction & Subjective & Ease of use \\
\hline
\end{tabularx}
\label{tab:xai-metrics}
\end{table}

\subsection{Evaluation of the Explanations}
In the literature presented above, there are over ten different evaluation metrics for the performance of the machine learning algorithms, including RMSE, MAPE, FP, etc. This shows that the field has collectively come to an agreement on how we should measure performance in a meaningful way. The evaluation of the explanations has not received the same attention as the performance of the algorithm even though work has been done in defining these different metrics, some of which are seen in Table \ref{tab:xai-metrics}.

Miller \cite{Miller2019-xc} provides one of the most in-depth descriptions of various people’s needs regarding explanations. Miller has provided many theoretical representations for explanation including scientific explanations and data explanations. They also provide much more information including levels of explanation that could be applicable to different types of users, structures of explanations that could impact the power of the explanations, and more. 

Coroama and Groza \cite{coroama2022evaluation} present 37 different metrics for measuring the effectiveness of an explanation. The methods range from objective to subjective types. Each method includes the property it measures and whether there is a systemic implementation. 

Sisk et al.\cite{sisk2022analyzing} present the case for human-centered evaluations and objective evaluations for explainable methods. Their human-centered evaluations aim at partitioning the users based on their wants from explainable systems. The objective metrics provided involve many aspects of the explanations including number of rules and number of features. 

Kadir et al.\cite{kadirassessing} propose a taxonomy of XAI evaluations as they appeared in the literature. They identified 28 different metrics through their literature search. These metrics are broken down into a taxonomy of how the analysis is performed. An example would be sensitivity analysis for local explanations. Sensitivity analysis is broken down into the removal of features and the addition of features. Each of these categories then includes many methods that were used. 

Hoffman et al.\cite{hoffman2018metrics} express the importance of high quality explanations in XAI. If explanations are received well and are valid, a user would be better equipped to trust and use a system that employes the XAI process. This allows for multiple areas of evaluations including the \emph{goodness} of the explanation, the \emph{satisfaction} the explanations provided to the users,  the \emph{comprehension} of the user, the \emph{curiosity} that motivates the user, the \emph{trust and reliance} the user has with the AI, and the \emph{performance} of the human-XAI system. They provide methods for measuring these metrics that are readily available.

\begin{figure}[!tbp]
  \centering
  \includegraphics[width=0.5\textwidth]{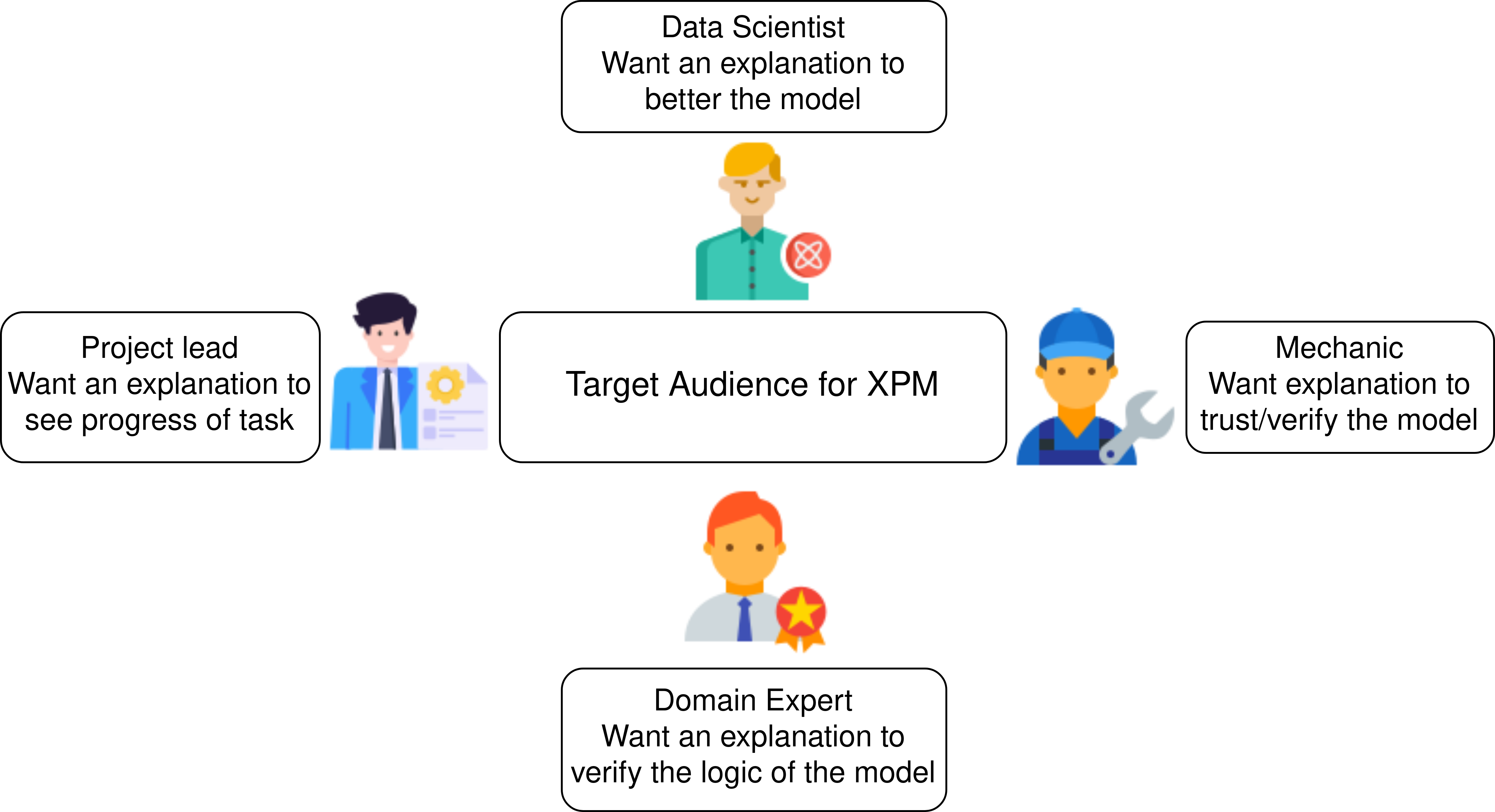}
    \caption[Potential Audiences of Explainable Predictive Maintenance. Icons taken from]{Potential Audiences of Explainable Predictive Maintenance. Icons taken from \footnotemark }
    \label{fig:xpm_audience}
\end{figure}
\footnotetext{https://icons8.com/}
\subsection{Addition of Human Involvement}
The target audience of an explainable system is a human subject whether a data scientist, a stakeholder, an engineer, or other. Addressing the needs of different types of users of an explainable system is an important area of research that is currently lacking. As seen in Fig. \ref{fig:xpm_audience}, different people on the same task have different goals and desires from predictive maintenance. While compensating for these differences would be difficult, we suggest a way to accomplish this, together with the resulting benefits.

First, a target audience for the explainable system should be identified, ensuring that a sample population of statistically significant size is used. Presenting the information to this sample population would bring many benefits to the XAI field as a whole. These include: making more quality metrics available, allowing researchers to discern which information is more or less useful, and bringing more attention to customizable explanations via the type of user. These would push the field of XAI forward as well as push the field of predictive maintenance forward towards a human-AI teaming environment. 

\subsection{Study Limitations}
This study focuses on a small amount of potential XAI and iML literature. While this survey reflects the work done as applied to predictive maintenance, it does not reflect many of the applied XAI and iML algorithms that exist. It also does not reflect all of the applicable ML algorithms developed within the context of predictive maintenance. While we do not see this as a detriment to the article presented, we do note that there are a number of popular methods of which the reader may be aware that are not present. 

\section{Conclusion}
\label{sec:conclusion}
Over the last decade, predictive maintenance has occupied a  considerable presence in the field of machine learning research. As we move towards complex mechanical systems with interdependencies that we struggle to explain, predictive maintenance allows us to break down the mysticality of what could potentially go wrong in the system. Many of these approaches move us closer to understanding the system while building a new system that we need to comprehend. Explainable predictive maintenance and interpretable predictive maintenance aim at breaking down these new walls to bring us closer to a clear understanding of the mechanical system. 

In this review, we provided a wide range of methods that are being used to tackle the problem of explainability. These methods are broken down in XAI and iML approaches. In our writing, XAI was broken-up into model-agnostic approaches like SHAP, LIME and LRP, and model-specific approaches like GradCAM and DIFFI. iML approaches all apply different methods of applying inherently interpretable models to the problem of predictive maintenance. 

Our systematic review of XAI and iML as applied to predictive maintenance showed some weak points in the field that can be addressed. Namely, there is a lack of utilization of metrics of explanations in predictive maintenance. The field of XAI has shown a number of metrics that do not even need to show the explanations to the target audience of the explainable systems. We provided a list of potential metrics found in the literature that can be applied to this domain. 

Lastly, we provided a short description of how humans can be brought into the evaluation of explainable and interpretable methods. After defining the target audience, researchers can gather a statistically significant sized sample of that audience. Providing the explanations to that sample would give feedback and allow the field to push towards human-specified explanations. 

\bibliographystyle{IEEEtran}
\bibliography{references}

\begin{IEEEbiography}[{\includegraphics[width=1in,height=1.25in,clip,keepaspectratio]{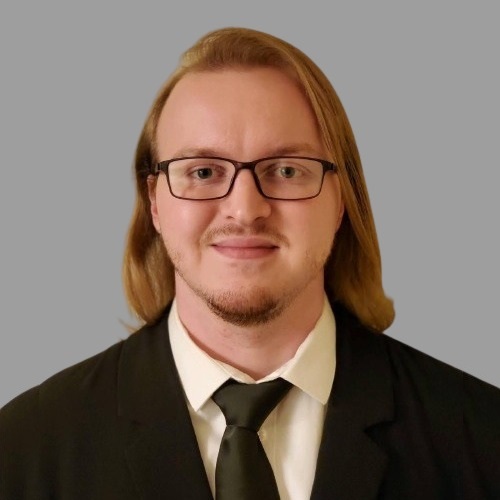}}]{Logan Cummins} (Member, IEEE) received their B.S. degree in Computer Science and Engineering from Mississippi State University. They are currently pursuing a Ph.D degree in Computer Science at Mississippi State University with a minor in Cognitive Science. 

They are a Graduate Research Assistant with the Predictive Analytics and Technology Integration (PATENT) Laboratory in collaboration with the Institute for Systems Engineering Research. Additionally, they perform research with the Social Therapeutic and Robotic Systems (STaRS) research lab. Their research interests include explainable artificial intelligence and its applications, cognitive science, and human-computer interactions as applied to human-agent teamming. They are a member of ACM and IEEE at Mississippi State University.
\end{IEEEbiography}

\begin{IEEEbiography}[{\includegraphics[width=1in,height=1.25in,clip,keepaspectratio]{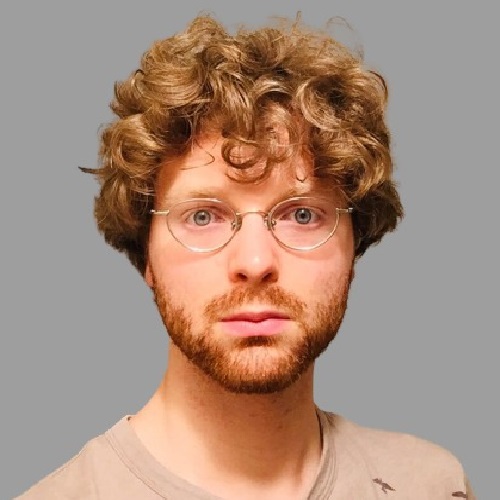}}]{Alexander Sommers} (Member, IEEE) received his B.S. in Computer Science from Saint Vincent Collage and his M.S. in Computer Science from Southern Illinois University. He is pursuing A Ph.D. in Computer Science at Mississippi State University with a concentration in machine learning.

He is a Graduate Research Assistant in the Predictive Analytics and Technology Integration Laboratory (PATENT Lab), in collaboration with the Institute for Systems Engineering Research. His work concerns synthetic time-series generation and remaining-useful-life prediction. His interests are the application of machine learning to reliability engineering and lacuna discovery respectively. He is a member of IEEE and ACM.
\end{IEEEbiography}

\begin{IEEEbiography}[{\includegraphics[width=1in,height=1.25in,clip,keepaspectratio]{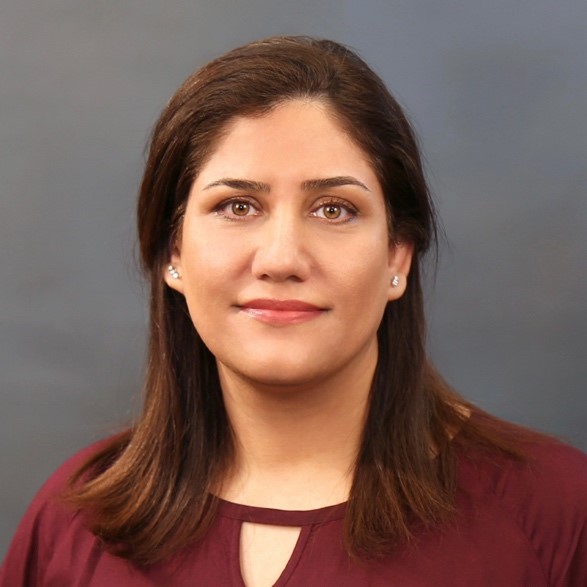}}]{Somayeh Bakhtiari Ramezani} (Member, IEEE) received the B.S. degree in computer engineering and the M.S. degree in information technology engineering from the Iran University of Science and Technology, in 2004 and 2008, respectively. She is currently pursuing the Ph.D. degree in computer science with Mississippi State University.

She is a Graduate Research Assistant with the Predictive Analytics and Technology Integration (PATENT) Laboratory in collaboration with the Institute for Systems Engineering Research. Prior to joining Mississippi State University, in 2019, she was with several companies in the energy and healthcare sectors as an HPC programmer and a data scientist. She is a 2021 SIGHPC Computational and Data Science Fellow. Her research interests include probabilistic modeling and optimization of dynamic systems, the application of ML, quantum computation, and time-series segmentation in the healthcare sector. She is a member of an ACM, the President of the ACM-W Student Chapter at Mississippi State University, and the Chair of the IEEE-WIE AG Mississippi section.
\end{IEEEbiography}

\begin{IEEEbiography}[{\includegraphics[width=1in,height=1.25in,clip,keepaspectratio]{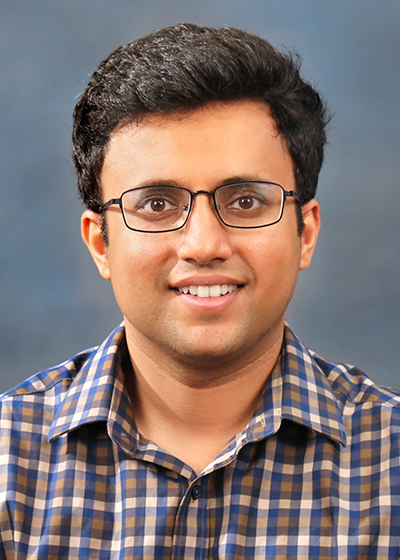}}]{Sudip Mittal} (Member, IEEE) is an Assistant Professor in the Department of Computer Science \& Engineering at the Mississippi State University. He graduated with a Ph.D. in Computer Science from the University of Maryland Baltimore County in 2019. His primary research interests are cybersecurity and artificial intelligence. Mittal’s goal is to develop the next generation of cyber defense systems that help protect various organizations and people. At Mississippi State, he leads the Secure and Trustworthy Cyberspace (SECRETS) Lab and has published over 80 journals and conference papers in leading cybersecurity and AI venues. Mittal has received funding from the NSF, USAF, USACE, and various other Department of Defense programs. He also serves as a Program Committee member or Program Chair of leading AI and cybersecurity conferences and workshops. Mittal’s work has been cited in the LA times, Business Insider, WIRED, the Cyberwire, and other venues.  He is a member of the ACM and IEEE.
\end{IEEEbiography}

\begin{IEEEbiography}
[{\includegraphics[width=1in,height=1.25in,clip,keepaspectratio]{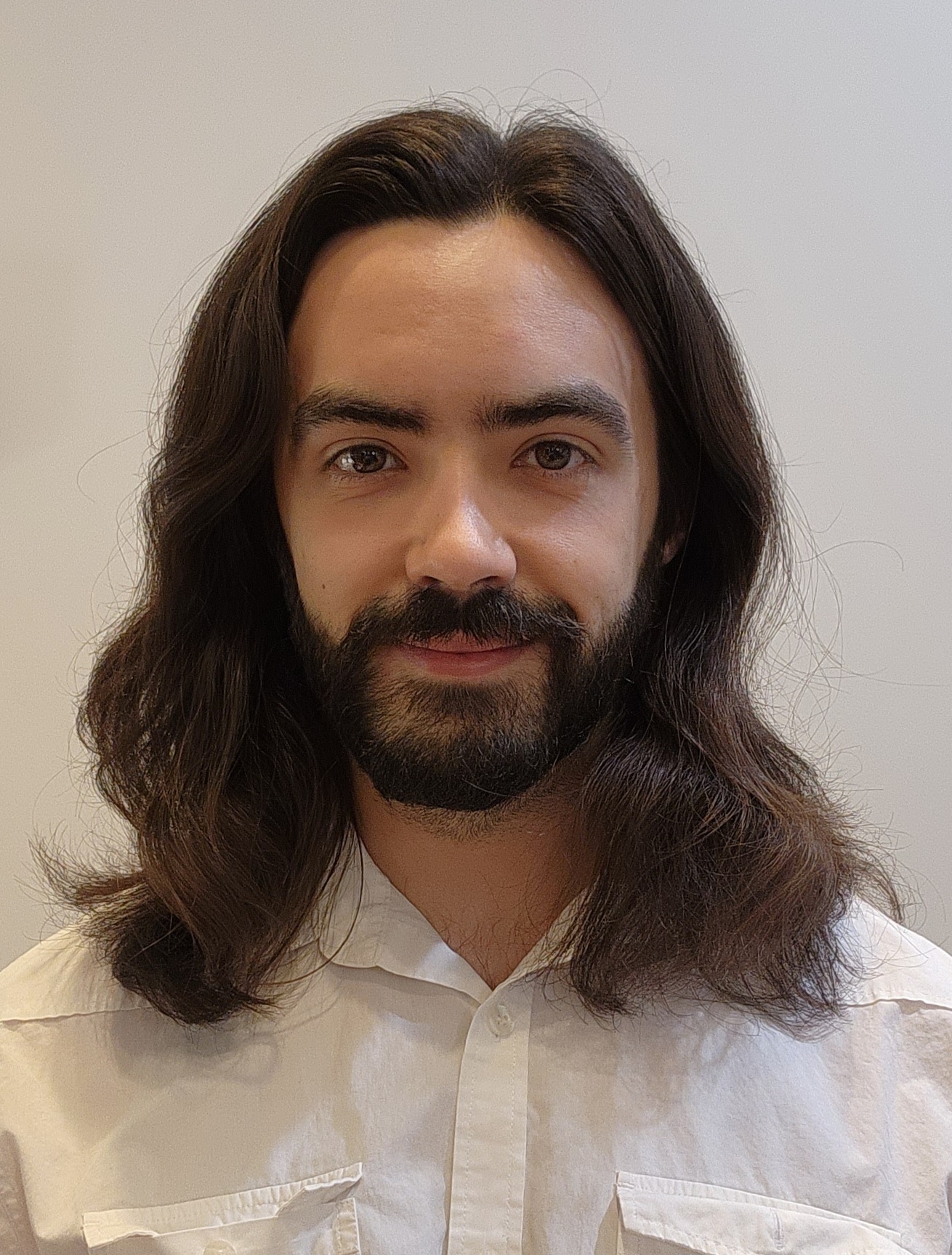}}]{Joseph Jabour} received his B. S. of Computer Science from the University of Mississippi in 2019, and is currently pursuing an M. S. of Computer Science from Mississippi State University.

He is a Computer Scientist at the Information Technology Lab (ITL) of the Engineering Research and Development Center (ERDC) in Vicksburg, MS. He began working at the ERDC in 2019, and he has pursued research in the field of Artificial Intelligence and Machine Learning. Additionally, he has performed a significant amount of work in the fields of Data Visualization, Digital Twins, and many other forms of research. He has since presented at several nationally recognized conferences, has held a vice chair position in the ERDC Association of Computing Machinery, and is currently a facilitator of the ERDC ITL Field Training Exercise based off of leadership principles from the Echelon Front. He has received awards for his research and development including but not limited to the ERDC Award for Outstanding Innovation in Research and Development. He seeks to push past the forefront of technological development and innovation and endeavors to identify and implement solutions to our nation's leading causes of concern.
\end{IEEEbiography}

\begin{IEEEbiography}
[{\includegraphics[width=1in,height=1.25in,clip,keepaspectratio]{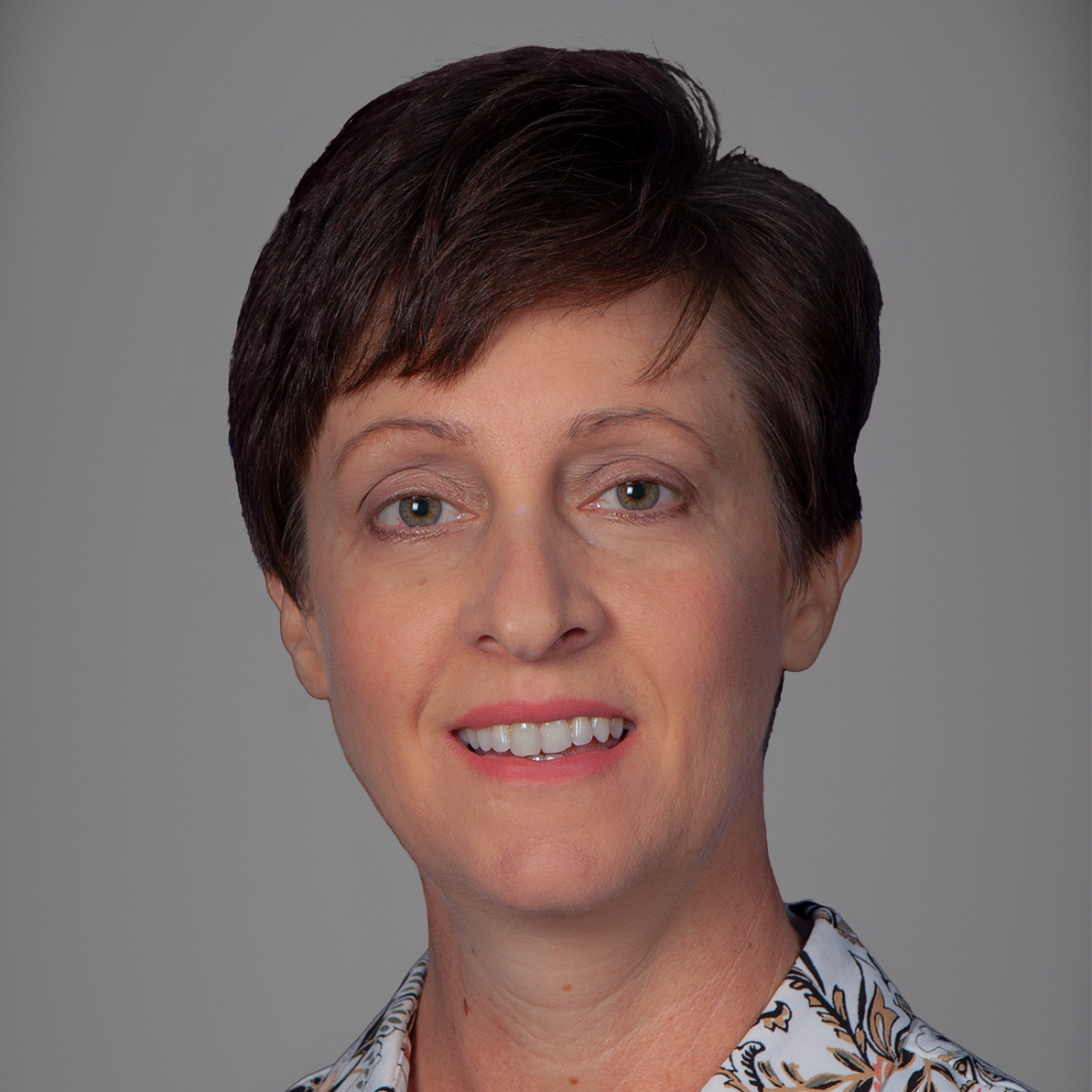}}]{Maria Seale} received the B.S. degree in computer science from the University of Southern Mississippi, in 1987, and the M.S. and Ph.D. degrees in computer science from Tulane University, in 1992 and 1995, respectively.

Prior to joining the Information Technology Laboratory, U.S. Army Engineer Research and Development Center (ERDC), in 2016, she held positions with the Institute for Naval Oceanography, the U.S. Naval Research Laboratory, and various private companies, as well as a tenured Associate Professorship with the University of Southern Mississippi. At ERDC, she has been involved with research in making scalable machine learning algorithms available on high-performance computing platforms and expanding the center’s capabilities to manage and analyze very large data sets. Her research interests include natural language processing, machine learning, natural computing, high-performance data analytics, and prognostics and health management for engineered systems. She is a member of the Prognostics and Health Management Society, the American Society of Mechanical Engineers, and the Association of Computing Machinery.
\end{IEEEbiography}

\begin{IEEEbiography}[{\includegraphics[width=1in,height=1.25in,clip,keepaspectratio]{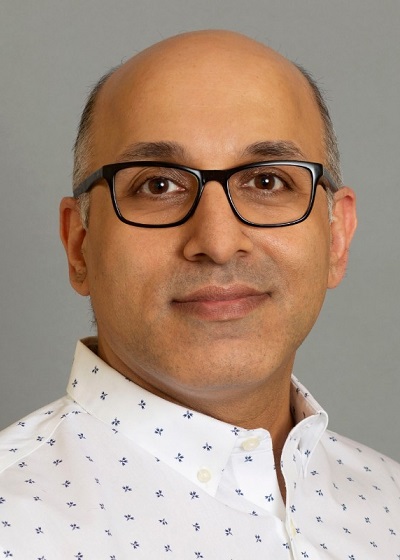}}]{Shahram Rahimi} (Member, IEEE) is currently a Professor and the Head of the Department of Computer Science and Engineering, Mississippi State University. Prior to that, he led the Department of Computer Science, Southern Illinois University, for five years. He is also a recognized leader in the area of artificial and computational intelligence, with over 220 peer-reviewed publications and a few patents or pending patents in this area.

He is a member of the IEEE New Standards Committee in Computational Intelligence. He provides advice to staff and administration at the federal government on predictive analytics for foreign policy. He was a recipient of the 2016 Illinois Rising Star Award from ISBA, selected among 100s of highly qualified candidates. His intelligent algorithm for patient flow optimization and hospital staffing is currently used in over 1000 emergency departments across the nation. He was named one of the top ten AI technology for healthcare, in 2018, by HealthTech Magazine. He has secured over \$20M of federal and industry funding as a PI or a co-PI in the last 20 years. He has also organized 15 conferences and workshops in the areas of computational intelligence and multi-agent systems over the past two decades. He has served as the Editor-in-Chief for two leading computational intelligence journals and is on the editorial board of several other journals.
\end{IEEEbiography}
\EOD
\end{document}